\documentclass[sn-mathphys]{sn-jnl}% Math and Physical Sciences Reference Style
\usepackage{graphicx}
\usepackage{multirow}
\usepackage{amsmath, amssymb}
\usepackage{adjustbox}
\usepackage{subfig}

\DeclareMathOperator{\E}{\mathbb{E}}
\DeclareMathOperator{\argmax}{arg\,max}
\DeclareMathOperator{\argmin}{arg\,min}
\jyear{2021}%

\theoremstyle{thmstyleone}%
\theoremstyle{thmstyletwo}%

\theoremstyle{thmstylethree}%

\begin{document}

\title[Balancing policy constraint and ensemble size in offline-RL]{Balancing policy constraint and ensemble size in uncertainty-based offline reinforcement learning}


\author[1,3]{ \fnm{Alex} \sur{Beeson} }\email{alex.beeson@warwick.ac.uk}
\author[2,3,4]{\fnm{Giovanni} \sur{Montana}}\email{g.montana@warwick.ac.uk\footnote{Corresponding author}}
\affil[1]{Warwick Medical School, University of Warwick, Coventry}
\affil[2]{Department of Statistics, University of Warwick, Coventry}
\affil[3]{WMG, University of Warwick, Coventry}
\affil[4]{Alan Turing Institute, London}


\abstract{Offline reinforcement learning agents seek optimal policies from fixed data sets.  With environmental interaction prohibited, agents face significant challenges in preventing errors in value estimates from compounding and subsequently causing the learning process to collapse.  Uncertainty estimation using ensembles compensates for this by penalising high-variance value estimates, allowing agents to learn robust policies based on data-driven actions.  However, the requirement for large ensembles to facilitate sufficient penalisation results in significant computational overhead.  In this work, we examine the role of policy constraints as a mechanism for regulating uncertainty, and the corresponding balance between level of constraint and ensemble size.  By incorporating behavioural cloning into policy updates, we show empirically that sufficient penalisation can be achieved with a much smaller ensemble size, substantially reducing computational demand while retaining state-of-the-art performance on benchmarking tasks.  Furthermore, we show how such an approach can facilitate stable online fine tuning, allowing for continued policy improvement while avoiding severe performance drops.}

\keywords{offline reinforcement learning, ensemble based uncertainty estimation, behavioural cloning, online fine-tuning, pessimism}


\maketitle

\section{Introduction}\label{Intro}
Reinforcement learning (RL) is concerned with optimising sequential decision-making in dynamic environments \citep{tesauro1995temporal, sutton2018reinforcement}. Typically, RL is used to train autonomous agents to perform complex tasks that rely on long-term decision making, where the decisions themselves impact future decisions as well as the environment the agent learns in. The agent identifies the optimal sequence of decisions, or actions, through trial-and-error learning, constantly interacting with the environment and adjusting its behaviour based on the rewards received. The end goal is to discover a policy that maximizes environmental rewards. By combining RL with the powerful predictive capabilities of neural networks, deep reinforcement learning has produced notable success in areas such as gaming \cite{mnih2013playing, hessel2018rainbow}, robotics \cite{kalashnikov2018qt, mahmood2018benchmarking} and autonomous driving \cite{RLAutoDriving}, advancing each year as it garners increasing interest and attention.   

Despite the remarkable achievements of RL, its reliance on continuous interaction with the environment restricts its application in areas where data collection is expensive, time-consuming, or hazardous. While simulators can partially alleviate this issue in fields such as robotics and autonomous driving \cite{todorov2012mujoco}, there are numerous situations where these are unavailable, and the trial-and-error nature of RL is clearly unsuitable or even unethical (e.g. in healthcare). Furthermore, these settings often already possess a wealth of data amassed through routine data collection or experimentation, offering a rich information source before an agent even engages in any environmental interaction \cite{komorowski2018artificial, liu2020reinforcement, yu2021reinforcement}.

The ambition to extend RL into such domains has given rise to offline reinforcement learning (offline-RL) \citep{lange2012batch}, a paradigm where agents are restricted from interacting with the environment and must learn exclusively from pre-existing interactions. Conventional RL algorithms typically falter in this offline setting, as the primary method for rectifying errors in action value estimates is no longer available.  This often leads to a complete collapse of the learning process as these errors propagate and compound during training \citep{fujimoto2019off}. Essentially, it is difficult for an agent to accurately assess the value of actions it has never encountered before, undermining the process of learning a policy based on value estimation.

The most common approach for overcoming this problem is to perform some kind of regularisation during training, encouraging updates during policy evaluation and/or policy improvement to stay close to actions in the underlying data \citep{levine2020offline}.  To date, numerous approaches have been proposed, ranging from methods that directly target the policy and/or value estimates \citep{kumar2019stabilizing, wu2019behavior, kumar2020conservative, nair2020awac, kostrikov2021offlineIQL, brandfonbrener2021offline} through to those which incorporate models of the environment \citep{kidambi2020morel, yu2021combo, argenson2020model, janner2022planning}, each with their own strengths and weaknesses in terms of performance, computational efficiency, reproducibility, hyperparameter optimisation and ease of implementation.

One such approach centres around uncertainty quantification with respect to the estimated value of actions \citep{abdar2021review}.  For actions absent in data, commonly referred to as out-of-distrubtion (OOD) actions, values estimates are subject to higher uncertainty than those present in data.  In online settings, this is often used to improve exploration by being optimistic in the face of uncertainty \cite{ciosek2019better,  chen2017ucb}.  Offline, this is used to stay closer to actions in the data by, conversely, being pessimistic in the face of uncertainty \cite{buckman2020importance}.  Specifically, action-value estimates are penalised based on their level of uncertainty, in effect guiding the agent towards actions that are high-value/low-variance.  

Although there are several techniques available for uncertainty quantification, ensemble-based methods in particular have found favour in offline-RL.  SAC-N \cite{an2021uncertainty}, for example, utilises an ensemble of value functions to approximate a value distribution, using the minimum value across the ensemble to penalise estimates pessimistically, attaining strong performance on offline benchmarks.  However, the ensemble size needed to realise this minimum can be excessively large, resulting in substantial computational overhead and scalability issues. While alternative approaches attempt to alleviate this by promoting greater diversification across the ensemble \cite{an2021uncertainty} or incorporating elements of conservative value estimation \cite{ghasemipour2022so}, they still remain relatively computationally demanding.

Recognising the potential of ensemble-based approaches to offline-RL, in this work we aim to address this practical obstacle through the use of policy constraints.  In offline-RL, policy constraints have been extensively employed as a method for ensuring OOD policy actions stay closer to data actions. Here, we investigate its role as a simple method for controlling the effective sample size of OOD actions, thus directly regulating the degree of epistemic uncertainty of value functions assessed for these actions.

Our findings indicate that when using unconstrained policies, the level of uncertainty in value estimates for OOD actions is relatively low, necessitating the use of large ensemble sizes to accurately estimate the tails of value distributions, and thus achieve the minimal values required for sufficient penalisation.  Using a constrained policy on the other hand, results in increased epistemic uncertainty, proportional to the strength of constraint and distance from data actions. Due to the heightened uncertainty, the tails of the value distribution become elongated, allowing for the acquisition of similar minimal values with a considerably reduced ensemble size. We find this to be the case when using two alternative methods for training the ensemble of value functions, namely shared and independent target values.

We leverage these findings as part of two distinct implementations based on existing offline-RL algorithms: TD3-BC-N (an extension of the TD3-BC \cite{fujimoto2021minimalist}) and SAC-BC-N (an extension of SAC-N).  In both cases, the policy constraint takes the form of behavioural cloning (BC), avoiding the need to explicitly model the behaviour of data actions, with inherent benefits in terms of simplicity and efficiency.  Moreover, we use BC to extend these approaches to online fine-tuning, gradually diminishing its influence as the agent interacts with the environment.

Through an extensive empirical evaluation using the D4RL benchmarking suite \citep{fu2020d4rl}, we show both implementations are able to produce state-of-the-art policies in a computationally efficient manner, which can then be fine-tuned during deployment while largely mitigating severe performance drops during the offline-to-online transition. In addition, we find this can be achieved without having to adjust hyperparameters based on data quality, an arguably necessary feature for real-world application where the performance properties of the data may be undetermined. We hope our work highlights the potential of such an approach and provides a useful benchmark for future advancements to be evaluated against.  For the purpose of transparency and reproducibility, the code base for this work is made freely available\footnote{https://github.com/AlexBeesonWarwick/OfflineRLConstrainedEnsemble}. 

The remainder of this manuscript is structured as follows.  In Section \ref{RelatedWork} we outline related work on behavioural cloning, uncertainty quantification and online fine-tuning before providing background material in Section \ref{Prelim}.  We present our offline learning and online fine-tuning procedures in Section \ref{PCUE} and evaluate them in Section \ref{Expts}.  We end with a discussion and concluding comments in Section \ref{DiscConc}.  

\section{Related work}\label{RelatedWork}
In this Section, we provide an overview of related literature on offline-RL and online fine-tuning. With respect to offline-RL, we focus on methods that utilise  behavioural cloning and uncertainty estimation as strategies to counteract overestimation bias for out-of-distribution actions. For online fine-tuning, we review methodologies that prioritize both stability and performance.

    \subsection*{Methods based on behavioural cloning}\label{RelatedWorkBC}
    In its most vanilla form, behavioural cloning (BC) is a form of imitation learning designed to mimic the actions of a demonstrator, most commonly an expert  \cite{bain1995framework}. Its use in offline-RL is primarily to act as a policy constraint, preventing agents from choosing actions that stray too far from the source data. 

    One way of incorporating BC into offline-RL is through modelling the distribution of actions in the data, commonly referred to as the behaviour policy.  In BCQ \citep{fujimoto2019off}, this is achieved using a Variational AutoEncoder (VAE) \citep{sohn2015learning}, whose generated actions form the basis of a policy which is then optimally perturbed by a separate network in the DDPG \cite{lillicrap2015continuous} framework.  This approach is modified by PLAS \citep{zhou2020plas} to train policies within the latent space of VAE, naturally constraining policies as they are decoded from latent to action space.  VAEs are also utilised by BRAC \cite{wu2019behavior} and BEAR \cite{kumar2019stabilizing}, which instead seek to minimise divergence metrics (Kullback-Leibler, Wasserstein, Maximum Mean Discrepancy) between the behaviour and the learned policy.  
    To account for multimodality, Fisher-BRC \citep{kostrikov2021offline} clones a behaviour policy using Gaussian mixtures and uses this for critic regularisation via the Fisher divergence metric.  Implicit Q-learning (IQL) \citep{kostrikov2021offlineIQL} combines expectile regression and advantaged weighted BC to train agents without having to evaluate actions outside the data.  TD3-BC \cite{fujimoto2021minimalist} favours a minimalist approach, directly incorporating BC into policy updates via a mean squared error between data and policy actions.
    
    Despite their diversity, each of these methodologies effectively addresses overestimation bias, facilitating the learning of a policy that either matches or surpasses the original behaviour. Additionally, they achieve this in a computationally efficient manner, requiring only a limited number of networks and relatively few gradient updates. However, these approaches tend to be overly restrictive, hindering agents' abilities to discern optimal behaviour from suboptimal data. Consequently, their performance is often inferior to alternative methods \cite{an2021uncertainty, ghasemipour2022so}.  Nonetheless, as we suggest, these techniques can still be employed in a complementary capacity alongside ensemble-based approaches, improving computational efficiency via fostering uncertainty for OOD value estimates.   
    
    \subsection*{Methods based on uncertainty quantification}\label{RelatedWorkUE}    
    As is customary in machine learning, we distinguish between two distinct sources of uncertainty: \textit{aleatoric} and \textit{epistemic} \citep{hullermeier2021aleatoric}.  The former stems from inherent stochasticity while the later arises due to incomplete information.  In deep learning, various techniques for quantifying both sources of uncertainty have been proposed (for extensive reviews see e.g. \citep{abdar2021review, zhou2022survey}) and several studies have endeavoured to provide insights in the context of RL (for instance \citep{eriksson2022sentinel, charpentier2022disentangling, lee2021sunrise}). These preliminary attempts have sought to address various challenges, including mitigating Q-learning instability, achieving equilibrium between exploration and exploitation, and facilitating risk-sensitive sequential decision-making.  
    
    In model-free RL, ensemble methods have garnered considerable interest for estimating epistemic uncertainty for action-value estimates.  In online-RL, ensembles are frequently employed to enhance exploration by encouraging agents to seek out actions whose estimated values vary the most.  This is achieved by constructing a distribution of action-value estimates using the ensemble and acting optimistically with respect to the upper bound, as demonstrated by \cite{ciosek2019better,  chen2017ucb}. In offline-RL these distributions direct agents towards actions within the dataset by, conversely, acting pessimistically with respect to the lower bound, prioritizing actions characterized by high value and low variance.
    
    SAC-N \cite{an2021uncertainty}, for example, adapts SAC \citep{haarnoja2018soft, haarnoja2018softauto} to offline setting by increasing the number of critics from $2$ to $N$, choosing the minimum across the ensemble to penalise action-value estimates that vary the most.  While very effective in term of performance, in some cases the size of the ensemble needed to estimate this minimum is excessively large (up to 500) as is the number of gradient steps required to reach peak performance (up to 3M).  Even with parallelisation, this results in considerable computational overhead, both in terms of training time and memory requirements, affecting the capacity to scale up to more complex, real-world problems.  
        
    EDAC \citep{an2021uncertainty} attempts to reduce ensemble size by increasing uncertainty through diversification. The authors note that, without intervention, the gradients of the critic ensemble tend to align, requiring larger and larger ensembles to achieve sufficient penalisation. To counteract this, EDAC diversifies these gradients by minimising the pair-wise cosine similarity within the ensemble, reducing its size by as much as a factor of ten without compromising performance.  However, this diversity regulariser can still be relatively expensive for medium-sized ensembles and the large number of gradient updates remain.  Our proposed solution is instead based on increasing uncertainty through the use of policy constraints.
        
    The approach most similar to our own is MSG \cite{ghasemipour2022so}, which also uses an ensemble of critics for uncertainty estimation, but uses conservative Q-learning (CQL) \citep{kumar2020conservative} to steer agents towards actions in the data instead of BC.  In effect, CQL ``pushes down'' on value estimates for out-of-distribution actions and ``pushes up'' for actions in the data.  MSG replaces the shared target of SAC-N/EDAC with independent targets to enforce pessimism, and when combined with CQL performs well on challenging benchmarks.  However, this performance is still dependent on relatively large ensembles and many gradient steps, with attempts to mitigate this using more efficient means such as multi-head \cite{lee2015m} and multi-input/multi-outputs \cite{havasi2020training} leading to detrimental impacts on performance.  In contrast, our proposed solution emphasises mitigation through the application of BC.
    
    \subsection*{Methods for online fine-tuning}\label{RelatedWorkFT}    
    Depending on the quality of the dataset, offline trained agents may exhibit limited performance upon deployment, necessitating further online fine-tuning through interaction with the environment. It can be argued that the domains which necessitate offline learning to begin with also necessitate a smooth transition from offline to online learning, that improvements in performance should not be preceded by periods of policy degradation.  In practice, this presents a formidable challenge due to the sudden distribution shift from offline to online data, which can introduce bootstrapping errors that distort the pre-trained policy \citep{lee2020addressing}. While continued regularisation can potentially mitigate this issue, it can also hinder the agent's ability to learn from newly acquired samples.  As such, approaches that promote stability as well as performance are desirable.

    An initial theoretical study of policy fine-tuning in episodic Markov Decision Processes in \citep{xie2021policy}, examines the potential benefits of granting online agents access to a reference policy that is, in a certain sense, already close to an optimal one.  The policy expansion scheme proposed in \citep{zhang2023policy} attempts to achieve stable learning by using offline-trained policies as potential candidates within a policy set, while  REDQ+AdaptiveBC \cite{zhao2021adaptive} seeks stability through adaptively adjusting the BC component of TD3-BC based on online returns.  We make use of a similar approach proposed by \citep{beeson2022improving}, which adjust the influence of BC based on exponential decay, avoiding the need for prior domain knowledge as required by REDQ+AdaptiveBC.

    Other related studies have investigated different setups or aspects, such as action-free offline datasets (i.e., datasets without logged actions) \citep{zhu2023guiding} or ``learning on the job' \cite{nair2022learning} to improve policy generalisation. The feasibility of employing existing off-policy methods to capitalize on offline data through minimal algorithmic adjustments has be examined in \cite{nair2022learning}.  Their findings underscore the significance of sampling mechanisms for offline data, the crucial role of normalizing the critic update, and the advantages of large ensembles for improving sample efficiency.

\section{Preliminaries}\label{Prelim}
In this section, we present the common RL setup and outline the challenges encountered when adapting algorithms to the offline setting. We then provide details of ensemble-based uncertainty methods we adopt as part of our approach.

    \subsection{Offline reinforcement learning}\label{PrelimRL}
    We follow standard convention and define a Markov decision process (MDP) with state space $S$, action space $A$, transition dynamics $T(s'\mid s, a)$, reward function $R(s, a)$ and discount factor $0 < \gamma \le 1$ \citep{sutton2018reinforcement}.  An agent interacts with this MDP by following a policy $\pi(a \mid s)$, which can be deterministic or stochastic.  The goal of reinforcement learning is to discover an optimal policy $\pi^*(a \mid s)$ that maximises the expected discounted sum of rewards, 
    $$
    \E_\pi\sum_{t=0}^\infty\gamma^t r(s_t, a_t),
    $$
    also know as the return. In actor-critic methods, this is achieved by alternating between policy evaluation and policy improvement using Q-functions $Q^\pi(s,a)$, which estimate the value of taking action $a$ in state $s$ following policy $\pi$ thereafter.  Policy evaluation consists of updating the Q-function (the critic) based on the Bellman expectation equation $$
    Q^\pi(s, a)=r(s, a) + \gamma \E_{s' \sim T, a' \sim \pi} (Q^\pi(s', a')),
    $$
    where $s'$ and $a'$ are used to denote the next state and next action, respectively.  Policy improvement comes in the form of updating the policy (the actor) so as to maximise $Q(s, a)$. 
    
    In terms of objective functions, policy evaluation and policy improvement are defined as, respectively, 
        \begin{equation}\label{PolicyEval}
        Q^\pi = \underset{Q}{\argmin} \E_{(s, a, s') \sim D} \Big( Q(s, a) - r(s, a) -\gamma Q^\pi(s', \pi(s')) \Big)^2 ,
    \end{equation}
    and
    \begin{equation}\label{PolicyImprov}
        \pi = \underset{\pi}{\argmax} \E_{s \sim D}\Big[Q(s, \pi(s)) \Big] ,
    \end{equation}
    where $r(s, a) + \gamma Q^\pi(s', \pi(s'))$ is commonly referred to as the target value.
     
    In practice, both actor and critic are parameterised functions, employing non-linear approximation methods such as neural networks.  Parameters are updated according to sampled based estimates, with the samples themselves coming from the agent's own interactions with the environment.  To improve data efficiency, these interactions are stored in a replay buffer which is constantly added to and sampled from during training.  To encourage sufficient exploration of the environment, a level of randomness is induced into online action selection, such as by adding noise if policies are deterministic or sampling if policies are stochastic.

    In offline reinforcement learning, also known as batch reinforcement learning \citep{lange2012batch}, the agent no longer has access to the environment and instead must learn solely from pre-existing interactions $D=(s_i, a_i, r_i, s'_i)$.  While it is possible to adapt existing algorithms to this setting by simply removing online interaction, in practice this often leads to highly sub-optimal policies or a complete collapse of the learning process. The primary cause of this is the propagation and compounding of overestimation bias for state-action pairs absent in $D$ \citep{levine2020offline}.  Such overestimation bias results from the bootstrapped nature of Q-network updates and the maximisation carried out as part of policy improvement.  
    
    This can be seen more clearly by examining the general objectives of policy evaluation and improvement. In policy evaluation (\ref{PolicyEval}), Q-value estimates for $Q(s,a)$ and $Q(s', a')$ use actions sampled from different policies, namely the behaviour policy $\pi_\beta(s)$ (i.e. the policy/policies that collected previous interactions), and the learned policy $\pi(s)$.  Errors that appear during policy evaluation propagate to policy improvement (\ref{PolicyImprov}), biasing actions that maximise spurious Q-values estimates.  This then feeds back into policy evaluation, compounding existing errors which then propagate to policy improvement, and so on.  In the online setting such bias can be mitigated by trialing policy actions in the environment, observing rewards and correcting Q-value estimates accordingly.  In the offline setting this is no longer permitted and hence additional measures must be implemented in order to stabilise training.

    \subsection{Regularisation through uncertainty estimation}\label{PrelimRegUE}    
    A sensible approach to combating overestimation bias is to target its root cause, namely the Q-values estimates themselves.  One tool for achieving this is uncertainty estimation, using the premise that Q-value estimates for out-of-distribution (OOD) actions are inherently more uncertain than for actions in the data.  This uncertainty can be used in training to favour Q-values with low-variance in policy evaluation and high-value/low-variance in policy improvement, in effect guiding the agent towards actions in the vicinity of the data.

    This idea forms the basis of approaches such as SAC-N and EDAC.  Both use an ensemble of $N$ Q-functions to approximate Q-value distributions, updating network parameters using the minimum across the ensemble for policy actions $\pi(s)$. In terms of the general objectives for policy evaluation and improvement, these become, respectively:
    \begin{equation}\label{PolicyEvalSACN}
        Q^\pi_i = \underset{Q}{\argmin} \E_{(s, a, s') \sim D} \Big( Q_i(s, a) - r(s, a) -\gamma \min_{i=1, ..., N}Q^\pi_i(s', \pi(s')) \Big)^2 ,
    \end{equation}
    and
    $$
        \pi = \underset{\pi}{\argmax} \E_{s \sim D}\Big[\min_{i=1, ..., N}Q_i(s, \pi(s)) \Big].
    $$
     Alternatively, as is done in MSG, each Q-function can be updated towards its own (rather than a shared) target value, giving a modified policy evaluation objective of:
    \begin{equation}\label{PolicyEvalMSG}
        Q^\pi_i = \underset{Q}{\argmin} \E_{(s, a, s') \sim D} \Big( Q_i(s, a) - r(s, a) -\gamma Q^\pi_i(s', \pi(s')) \Big)^2 .
    \end{equation}

    Using uncertainty estimation in this way constitutes a pessimistic approach to offline-RL.  By using the minimum across the ensemble, Q-value estimates for OOD actions are penalised according to their level of uncertainty.  By increasing the size of the ensemble, the minimum is realised more accurately, and hence with large enough $N$ the level of penalisation is sufficient to prevent overestimation bias.  In practice, such approaches attain strong performance, but the size of the ensemble required to accurately estimate this minimum is often very large, necessitating the use of considerable computational resource to implement.
    
\section{Policy constrained critic ensembles}\label{PCUE}
The key issue we seek to address in this work is the high computational cost of ensemble-based approaches to offline reinforcement learning, approaches that are otherwise very effective due to their strong performance and straightforward implementation.  These costs primarily stem from the need to use large ensembles to obtain accurate estimates of lower bounds, which form the basis of penalties applied to Q-value estimates for OOD actions.

As demonstrated by \citep{an2021uncertainty}, the strength of these penalties depend on both the size of the ensemble and the magnitude of the standard deviation.  Using the same example for illustrative purposes (itself based on \citep{royston1982expected}), if $Q(s, a)$ follows a Gaussian distribution with mean $\mu(s, a)$ and standard deviation $\sigma(s, a)$, the approximate expected minimum of a set of $N$ realisations is given by:
\begin{equation}\label{ExpectedMinGaus}
        \E \Big[ \underset{j=1, ..., N}{\min} Q_j(s, a) \Big] \approx \mu(s, a) - \Phi^{-1} \biggl( \frac{N-\frac{\pi}{8}}{N - \frac{\pi}{4} + 1} \biggl) \sigma(s, a),
    \end{equation}
where $\Phi$ is the cumulative distribution function of the standard Gaussian.

In general the distribution of $Q(s, a)$ is unknown, but the same basic principles apply.  In SAC-N, the size of the ensemble needed to sufficiently penalise Q-value estimates is high, as the standard deviation across the ensemble (i.e. level of uncertainty) is relatively small.  In order to achieve similar levels of penalisation with a reduced ensemble size, the level of uncertainty across the ensemble must be increased.  In EDAC this is achieved by diversifying the ensemble and in MSG by using conservative Q-learning.

Our proposed method for increasing this uncertainty is based on policy constraints.  We note that, although policy constraints are primarily used to steer agents towards actions in the data, this also has an effect on the level of uncertainty of Q-values estimates of OOD actions.  By constraining the policy, the Q-ensemble is trained on actions closer to the data, in effect reducing the effective sample size of OOD actions, which in turn increases epistemic uncertainty with respect to their Q-value estimates.  The higher the level of constraint, the greater the level of uncertainty as the tails of the value distribution expand.  Thus, policy constraints provide an additional mechanism for controlling uncertainty in Q-value estimates, which can be used to achieve sufficient levels of penalisation with a much reduced ensemble size.

With this in mind, we modify existing ensemble-based approaches to directly incorporate behavioural cloning into policy updates, in a similar vein to TD3-BC \cite{fujimoto2021minimalist}.  While many other approaches for constraining policies exist (see Section \ref{RelatedWork}), we favour this one in particular as it requires no explicit modelling of the behaviour policy $\pi_\beta$ and is straightforward to implement, computationally cheap, flexible enough to accommodate deterministic and stochastic policies and requires no changes to policy evaluation using either shared (\ref{PolicyEvalSACN}) or independent (\ref{PolicyEvalMSG}) targets.

Let $\rho(a)$ be a function representing a divergence metric between policy and data actions $a$.  The general policy improvement objective becomes:
\begin{equation}\label{PolImpRho}
    \pi = \underset{\pi}{\argmax} ~ \E_{(s, a) \sim D} \Big[\min_{i=1, ..., N}Q_i(s, \pi(s)) - ~ \beta \rho(a) \Big].
\end{equation}

The hyperparameter $\beta$ controls the balance between RL and BC, and by extension the level of uncertainty in Q-value estimate for OOD actions.  Lower values favour RL but also lead to lower levels of uncertainty.  Higher values increase uncertainty, but tip the balance towards BC, making it more difficult for the agent to discover high-value actions that lie beyond the data.  Thus, the aim is to find a value of $\beta$ that induces enough uncertainty without being too restrictive, allowing sufficient penalisation of Q-value estimates using a smaller ensemble.

Regardless of the form of $\rho(a)$, the balance in (\ref{PolImpRho}) is highly sensitive to Q-value estimates, which scale with rewards and vary across tasks.  Therefore, to keep this balance in check, following the example of TD3-BC we normalise estimates by dividing by the mean of the absolute values, such that:
\[
    Q_{norm}(s, \pi(s)) = \frac{Q(s, \pi(s))}{\E_{s \sim D} \mid Q(s, \pi(s)) \mid}.
\]

So far we have presented our approach within the general actor-critic framework, outlining the changes to policy evaluation and policy improvement from incorporating ensemble methods and behavioural cloning.  In Sections \ref{PCUE+TD3NBC} and \ref{PCUE+SACNBC} we present two specific versions based on TD3 \cite{fujimoto2018addressing} and SAC \cite{haarnoja2018softauto}, respectively, which are then evaluated in Section \ref{Expts} alongside our fine-tuning approach detailed in Section \ref{OnlineFT}.

    \subsection{TD3-BC-N}\label{PCUE+TD3NBC}
    Twin Delayed Deep Deterministic Policy Gradient (TD3) is an approach to reinforcement learning that proposes a number of techniques for addressing function approximation error in actor-critic methods, most notably DDPG.  Based on a deterministic policy, TD3 makes use of a dual critic network for policy evaluation and updates Q-functions and policies at a ratio of 2:1.  As is common with Q-learning approaches, target networks are used to stabilise training, both in policy evaluation and policy improvement. Exploration comes in the form of noise sampled from a Gaussian distribution.  

    We modify the baseline TD3 algorithm by increasing the number of critics from $2$ to $N$ and adding a BC term to policy updates in the form of a mean squared error (similar to TD3-BC).  Corresponding parameter updates and notation are as follows. Let $\theta_{i}$ and $\theta_{i}^{'}$ represent the parameters of the $i^{th}$ Q-network and target Q-network, respectively, and $\phi$ and $\phi'$ represent the parameters for a policy network and target policy network, respectively.  Let $\beta$ represent the BC coefficient, $N$ the ensemble size, $\tau$ the target network update rate, $\epsilon$ policy noise and $B$ a sample of transitions from dataset $D$.

    Each Q-network update is performed through gradient descent. For shared target values, we use:
    \begin{equation}\label{TD3-BC-N-PolEvalDep}
        \nabla_{\theta_i}\frac{1}{\vert B \vert} \sum_{(s, a, r, s') \sim B} \Big(Q_{\theta_i}(s, a) - r - \gamma \min_{i=1, ..., N}Q_{\theta_i}(s', a')\Big)^2,
    \end{equation}
    and for individual target values:
    \begin{equation}\label{TD3-BC-N-PolEvalIndep}
        \nabla_{\theta_i}\frac{1}{\vert B \vert} \sum_{(s, a, r, s') \sim B} \Big(Q_{\theta_i}(s, a) - r - \gamma Q_{\theta_i}(s', a')\Big)^2.
    \end{equation}
    In either case $a'=(\pi_{\phi'}(s') + $noise) with noise sampled from an $N(0, \epsilon)$ distribution.  The policy network update is performed through gradient ascent using:
    \begin{equation}\label{TD3-BC-N-PolImp}
        \nabla_{\phi}\frac{1}{\vert B \vert} \sum_{(s, a) \sim B} \min_{i=1, ..., N}Q_{\theta_i}\big(s, \pi_\phi(s)\big) - \beta \big(\pi_\phi(s) - a \big)^2.
    \end{equation}
    Target networks are updated using Polyak averaging:
    \begin{equation}\label{TD3-BC-N-Targets}
    \begin{split}
        \theta^{'}_i \leftarrow \tau \theta_i + (1 - \tau) \theta^{'}_i \\
        \phi^{'} \leftarrow \tau \phi + (1 - \tau) \phi^{'}.
    \end{split}
    \end{equation}
    The final procedure is presented in Algorithm \ref{alg:td3-bc-n}.
    
    \begin{algorithm}[h]
    \caption{TD3-BC-N}\label{alg:TD3-BC-N}
    \begin{algorithmic}
    \Require Behavioural cloning coefficient $\beta$, ensemble size $N$, discount factor $\gamma$, policy noise $\epsilon$, target network update rate $\tau$ and data set $D$
    \State Initialise critic parameters $\theta_i$, policy parameters $\phi$ and corresponding target parameters $\theta'_i$, $\phi'$.
    \For {$j=0$ to $J$}
       \State Sample minibatch of transitions $(s, a, r, s')$ from $D$
       \State Update Q-function parameters $\theta_i$ using equation (\ref{TD3-BC-N-PolEvalDep}) or (\ref{TD3-BC-N-PolEvalIndep})
       \State Update policy parameters $\phi$ using equation (\ref{TD3-BC-N-PolImp})
       \State Update target network parameters $\theta'_i$ using equation (\ref{TD3-BC-N-Targets})
    \EndFor
    \end{algorithmic}
    \end{algorithm}
    
    \subsection{SAC-BC-N}\label{PCUE+SACNBC}
    Soft Actor-Critic (SAC) is a maximum entropy approach to reinforcement learning.  Based on a stochastic policy, SAC augments the standard policy evaluation and improvement objectives of actor-critic methods with an entropy regulariser, in effect encouraging agents to maximise returns while acting as randomly as possible.  This helps boost exploration, which comes in the form of sampling actions from the policy.  Like TD3, SAC uses a dual critic with target networks to promote stability during policy evaluation, but forgoes a target network for policy improvement and uses a critic to actor update ratio of 1:1.
    
    We modify the baseline SAC algorithm by increasing the number of critics from $2$ to $N$ and by adding a BC term to policy updates.  Since the policy is stochastic, this BC term can take the form of either a mean-squared error or log-likelihood.  Corresponding parameter updates and notation are as follows. Let $\theta_{i}$ and $\theta_{i}^{'}$ represent the parameters of the $i^{th}$ Q-network and target Q-network, respectively, and $\phi$ represent the parameters for a policy network.  Let $\alpha$ represent the entropy coefficient, $\mathcal{H}$ the minimum entropy, $\beta$ the BC coefficient, $N$ the ensemble size, $\tau$ the target network update rate and $B$ a sample of transitions from dataset $D$. 
    
    Each Q-network update is performed through gradient descent. For shared target values we use:
    \begin{equation}\label{SAC-BC-N-PolEvalDep}
      \nabla_{\theta_i} \frac{1}{\vert B \vert}   \sum_{\substack{(s, a, r, s') \sim B \\ a' \sim \pi_\phi(s')}}  \Big( Q_{\theta_i}(s, a) - r - \gamma \min_{i=1, ..., N} Q_{\theta_i}(s', a') + \alpha \log \pi_\phi(a' \mid s') \Big)^2,
    \end{equation}
    and for individual target values:
    \begin{equation}\label{SAC-BC-N-PolEvalIndep}
        \nabla_{\theta_i} \frac{1}{\vert B \vert} \sum_{\substack{(s, a, r, s') \sim B \\ a' \sim \pi_\phi(s')}} \Big(Q_{\theta_i}(s, a) - r - \gamma Q_{\theta_i}(s', a') + \alpha \log \pi_\phi(a' \mid s') \Big)^2.
    \end{equation}
    The policy network update is performed through gradient ascent. For mean-squared error we use:
    \begin{equation}\label{SAC-BC-N-PolImpMSE}
        \nabla_{\phi}\frac{1}{\vert B \vert} \sum_{\substack{(s, a) \sim B \\ a_p \sim \pi_\phi(s)}} \min_{i=1, ..., N}Q_{\theta_i}\big(s, a_p \big) - \alpha \log \pi_\phi(a_p \mid s) - \beta \big(\pi_\phi(s) - a \big)^2.
    \end{equation}
    and for log-likelihood:
    \begin{equation}\label{SAC-BC-N-PolImpLL}
        \nabla_{\phi}\frac{1}{\vert B \vert} \sum_{\substack{(s, a) \sim B \\ a_p \sim \pi_\phi(s)}} \min_{i=1, ..., N}Q_{\theta_i}\big(s, a_p \big) - \alpha \log \pi_\phi(a_p \mid s) + \beta \log \pi_\phi (a \mid s).
    \end{equation}
    The entropy coefficient update is performed through gradient ascent using:
    \begin{equation}\label{SAC-BC-N-Ent}
        \nabla_{\alpha} \frac{1}{\vert B \vert} \sum_{\substack{s \sim B \\ a_p \sim \\pi_\phi(s)}} \alpha \Big(\log \pi_\phi (a_p \mid s) + \mathcal{H} \Big).
    \end{equation}
    Target networks are updated using Polyak averaging:
    \begin{equation}\label{SAC-BC-N-Targets}
        \theta^{'}_i \leftarrow \tau \theta_i + (1 - \tau) \theta^{'}_i
    . \end{equation}
    The final procedure is presented in Algorithm \ref{alg:sac-bc-n}.
    
    \begin{algorithm}[h]
    \caption{SAC-BC-N}\label{alg:sac-bc-n}
    \begin{algorithmic}
    \Require Behavioural cloning coefficient $\beta$, ensemble size $N$, discount factor $\gamma$, minimum entropy $\mathcal{H}$, target network update rate $\tau$ and data set $D$
    \State Initialise critic parameters $\theta_i$ and corresponding target parameters $\theta'_i$.  Initialise policy parameters $\phi$ and entropy coefficient $\alpha$
    \For {$j=0$ to $J$}
       \State Sample minibatch of transitions $(s, a, r, s')$ from $D$
       \State Update Q-function parameters $\theta_i$ using equation (\ref{SAC-BC-N-PolEvalDep}) or (\ref{SAC-BC-N-PolEvalIndep})
       \State Update policy parameters $\phi$ using equation (\ref{SAC-BC-N-PolImpMSE}) or (\ref{SAC-BC-N-PolImpLL})
       \State Update entropy parameter $\alpha$ using equation (\ref{SAC-BC-N-Ent})
       \State Update target network parameters $\theta'_i$ using equation (\ref{SAC-BC-N-Targets})
    \EndFor
    \end{algorithmic}
    \end{algorithm}

\subsection{Stable online fine-tuning}\label{OnlineFT}    
The main goal in offline-RL is to discover optimal behavioural from existing data sets, allowing agents to learn effective policies before being deployment in the environment.  Following deployment however, agents can collect more information about the environment, presenting opportunities for continued improvement via online fine-tuning.  As agents can now correct for value estimates through online interaction, it may seem natural to remove constraints imposed during offline learning, but in practice this can often result in an initial phase of policy degradation due to the abrupt transition from constrained to unconstrained learning (see Section \ref{RelatedWork}). In many situations, such degradation is deemed undesirable, emphasising the need for approaches that prioritize stability alongside performance.
    
During the transition from offline to online learning, an agent's policy should exhibit consistent improvement, surpassing its offline performance without experiencing periods of substantial deterioration. Our approach is well-suited to accomplishing these objectives. First, by making minimal modifications to existing algorithms, we largely preserve the core characteristics that contribute to their success online. Second, our utilisation of BC offers a convenient mechanism for stabilizing the transition by gradually reducing its influence over time. Numerous methods can achieve this, but for simplicitly we adopt an approach based on exponential decay as in \citep{beeson2022improving}.  Let $\beta_{start}$ and $\beta_{end}$ be the initial and final values of the BC component $\beta$, respectively, and $S$ the number of decay steps.  The exponential decay rate $\kappa_\beta$ is given by:
        \begin{equation}\label{beta_decay}
            \kappa_{\beta}=\exp\big[\frac{1}{S}\log\big(\frac{\beta_{end}}{\beta_{start}}\big)\big] .
        \end{equation}
    
Determining the appropriate use of existing data is also an important aspect of online fine-tuning. One option is to supplement the existing data with new transitions, enabling a seamless transition as the agent gradually acquires new information online. However, if the original data is sub-optimal, the online fine-tuning process may be slow, as the agent's offline-trained policy is not fully utilised. Alternatively, discarding the data allows the agent to improve its policy without being hampered by data it has already improved upon.  However, this could compromise stability in the initial stages due to limited experience and a paucity of data. We propose an approach that strikes a balance, adding new transitions to a portion of the original data before training.  We outline this fine-tuning procedure using TD3-BC-N in Algorithm \ref{algoTD3-BC-N-FT}.  The corresponding procedure for SAC-BC-N is provided in the Appendix.

    \begin{algorithm}[h]
       \caption{Online fine-tuning (TD3-BC-N)}\label{algoTD3-BC-N-FT}
    \begin{algorithmic}
       \State {\bfseries Require:} Ensemble size $N$, discount factor $\gamma$, policy variance $\epsilon$, target network update rate $\tau$, data set $D$, exploration noise $\sigma$ and decay parameters $\beta_{start}, \beta_{end}, S$
       \State Initialise pre-trained critic parameters $\theta_i$, policy parameters $\phi$ and corresponding target parameters $\theta'_i$, $\phi'$.
       \State Initialise environment and replay buffer $R$
       \State Populate $R$ with a proportion of transitions from $D$.
        \For {$k=0$ to $K$}
           \State Act in environment with exploration, $a \sim \pi_\phi(s) + N(0, \sigma)$
           \State Store resulting transition $(s, a, r, s')$ in $R$
        \EndFor
       \State Set decay rate $\kappa_{\beta}$ as per equation (\ref{beta_decay})
       \State Set $\beta = \beta_{start}$
       \For {$j=0$ to $J$}
           \State Act in environment with exploration, $a \sim \pi_\phi(s) + N(0, \sigma)$
           \State Store resulting transition $(s, a, r, s')$ in $R$
           \State Sample minibatch of transitions $(s, a, r, s')$ from $R$
            \State Update Q-function parameters $\theta_i$ using equation (\ref{TD3-BC-N-PolEvalDep}) or (\ref{TD3-BC-N-PolEvalIndep})
            \State Update policy parameters $\phi$ using equation (\ref{TD3-BC-N-PolImp})
           \State Update target network parameters $\theta'_i$ using equation (\ref{TD3-BC-N-Targets})
           \State Update BC coefficient $\beta = \max(\beta_{end}, \kappa_{\beta}\beta$)
       \EndFor
    \end{algorithmic}
    %\label{alg:td3-bc-n}
    \end{algorithm}

\section{Experimental results}\label{Expts}
In this section, we present a comprehensive evaluation of our offline learning and online fine-tuning procedures using the open-source D4RL benchmarking suite. Section \ref{Expts_D4RL} provides an overview of this benchmark and the domains we consider, with Section \ref{implementation} outlining implementation details.  In Section \ref{interpretation} we investigate our claims regarding the impact of policy constraints on uncertainty estimation, and examine the trade-off between ensemble size and level of constraint.  This is followed by a comparison of performance and computational efficiency in Section \ref{performance}, as well as a number of supplementary experiments to highlight the importance of individual components and implementation choices.  We end in Section \ref{Expts_OnlineFT} with an assessment of our fine-tuning strategy.

    \subsection{Benchmark datasets}\label{Expts_D4RL}    
    D4RL is a popular resource for benchmarking offline reinforcement learning algorithms.  The suite contains a wide range of tasks and data sets designed to test an agent's ability to learn effective policies in various settings.  We outline the domains considered in this work, and refer the reader to the original paper for further details \citep{fu2020d4rl}.

    \begin{itemize}

    \item{{\bf MuJoCo}.}
    This setting makes use of the hopper, halfcheetah and walker2d environments of the MuJoCo physics simulator \citep{todorov2012mujoco}, assessing how well agents learn from sub-optimal and/or narrow data distributions.  Each environment has four associated data sets: ``expert'' which contains transitions collected from an agent trained to expert level using SAC; ``medium'' which contains transitions collected from an agent trained to 1/3 expert level using SAC; ``medium-replay'' which contains the transitions used to train the medium-level agent; ``medium-expert'' which contains the combined transitions from ``medium'' and ``expert''.  In general this setting is considered one of the easier among the benchmark, with environments having well defined rewards structures and data sets comprising a decent proportion of near-optimal trajectories.

    \item{{\bf Maze2D}.}
    This settings involves moving a force actuated ball to a fixed target location.  Data is collected via a controller which starts and ends at random goal locations.  The purpose of this setting is to test an agent's ability to stitch together previous trajectories to reach the evaluation goal.  There are three increasingly difficult mazes: ``umaze'', ``medium'' and ``large''.  We focus on the more challenging sparse reward setting, in which the agent receives a reward of 1 when within a 0.5 unit radius of the target goal and 0 otherwise.

    \item{{\bf AntMaze}.}
    This setting replaces the ball from Maze2d with an more complex Ant robot, with episodes terminating once the Ant reaches the goal location.  Data is collected via a controller using two different methods: ``play'' in which the controller moves from hand-picked starting locations to hand-picked goals; ``diverse' in which the controller moves from random starting locations to random goals.  This setting is considered one of the more challenging as agents must learn to both control the Ant and stitch trajectories together using only sparse rewards.
    
    \item{{\bf Adroit}.}
    This setting makes use of the Adroit environment, controlling a high-dimensional robotic hand to perform specifics tasks.  The aim is to assess whether agents can learn from narrow data distributions (``cloned'') and human demonstrations (``human'') with sparse rewards.  We focus on the ``pen'' task as, similar to other approaches, this is the only task in which notable performance is achieved (see Appendix).

    \end{itemize}
    
    \subsection{Implementation details}\label{implementation}   
    Following the protocol of D4RL, we train agents using offline data sets and evaluate their performance in the simulated environment.  Performance is measured in terms of normalised score, with 0 and 100 representing random and expert policies, respectively.  Each experiment is repeated across five random seeds with reported results the mean normalised score $\pm$ one standard error across 50 evaluations for MuJoCo and 500 evaluations for Maze2d/AntMaze/Adroit (10 and 100 evaluations per seed, respectively).
    
    For both TD3-BC-N and SAC-BC-N each Q-network comprises a 3-layer MLP with ReLU activation functions and 256 nodes, taking as input a state-action pair and outputting a Q-value.  For TD3-BC-N the policy network comprises a 3-layer MLP with ReLU activation functions and 256 nodes, taking as input a state and outputting an action bound to [-1, 1] via tanh transformation.  For SAC-BC-N the policy network comprises the same architecture but instead outputs the mean and standard deviation of a Gaussian distribution which is also bound to [-1, 1] via tanh transformation.  Each approach retains the hyperparameters values of their online counterpart (full details are provided in the Appendix).

    Across all data sets, we train agents for 1M gradient steps using an ensemble size of $N=10$.  To help stabilise training for narrow data distributions, we inflate the value of the BC coefficient $\beta$ by a factor of 10 for the first 50k gradient steps.  We use shared targets for MuJoCo and Maze2d tasks and independent targets for AntMaze and Adroit.  We investigate the impact of each of these designs decisions as part of our ablations studies.

    For the BC component, we find the characteristics of each environment necessitate varying intensities, and for SAC-BC-N dictate its form (mean-squared error or log-likelihood).  We therefore adjust its intensity and/or form based on task type, but to better reflect real-world scenarios where the quality of the data is often unknown, we prohibit adjustments within the same task.  Values for each task and data set are provided in the Appendix.

     \subsection{The impact of policy constraints on uncertainty} \label{interpretation}
     Before we consider the full range of tasks and data sets, we first investigate the claims made in previous sections relating to the impact of policy constraints on uncertainty levels in Q-value estimates for OOD actions.  To do this, we train a number of agents using TD3-BC-N with dependent target values across a range of $N$ and $\beta$ on the ``hopper-medium-expert'' dataset, and examine the performance of resulting policies and uncertainty of Q-values estimates from resulting ensembles.

\begin{figure}[h] % h=here, t=top, b=bottom, p=new page, !=other preferences
            \begin{center}
                \includegraphics[width=0.8\textwidth]{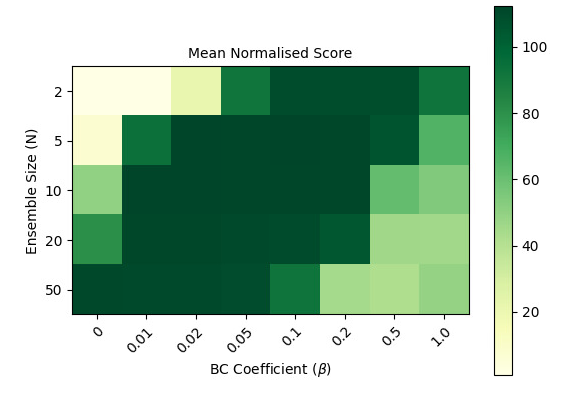}
            \end{center}
            \caption{Performance as a function of $N$ and $\beta$.  Lower values of $\beta$ require larger values of $N$ and smaller values of $N$ require higher values of $\beta$.  If both $N$ and $\beta$ are large, the uncertainty in Q-value estimates for OOD actions is too high, and thus the penalty applied too severe, leading the agent to prefer actions similar to those of the data}
            \label{fig:EnsVsBeta}
        \end{figure}

     Beginning with performance, we summarise this via a heatmap in Figure \ref{fig:EnsVsBeta}, using shade to represent mean normalised score.  For the lowest values of $\beta$ we see that larger ensembles are required to prevent overestimation bias through sufficient penalisation of OOD actions.  As the value of $\beta$ increases, the size of the ensemble required to achieve this level of penalty decreases, allowing the same level of performance to be attained as for larger ensembles.  We also see that the larger the value of $N$, the smaller the value of $\beta$ before performance starts to degrade.  In these cases, the level of uncertainty resulting from both a large ensemble and high level of policy constraint leads to over-penalisation of Q-values estimates, in effect driving the agent towards actions in the data at an increased rate.

    In terms of uncertainty of Q-value estimates, we consider both the standard deviation across the ensemble and the clip penalty $Q_{clip}(s, a)$, which measures the size of the difference between the mean and minimum:
    \[
    Q_{clip}(s, a) = \frac{1}{N} \sum_{j=1}^{N} Q(s, a) - \min_{j=1, ..., N}Q(s, a).
    \]
    In particular, we examine how each of these measures of uncertainty varies according to how far actions are from the data and the values of $N$ and $\beta$.

\begin{figure}[h] % h=here, t=top, b=bottom, p=new page, !=other preferences
            \begin{center}
                \includegraphics[width=1\textwidth]{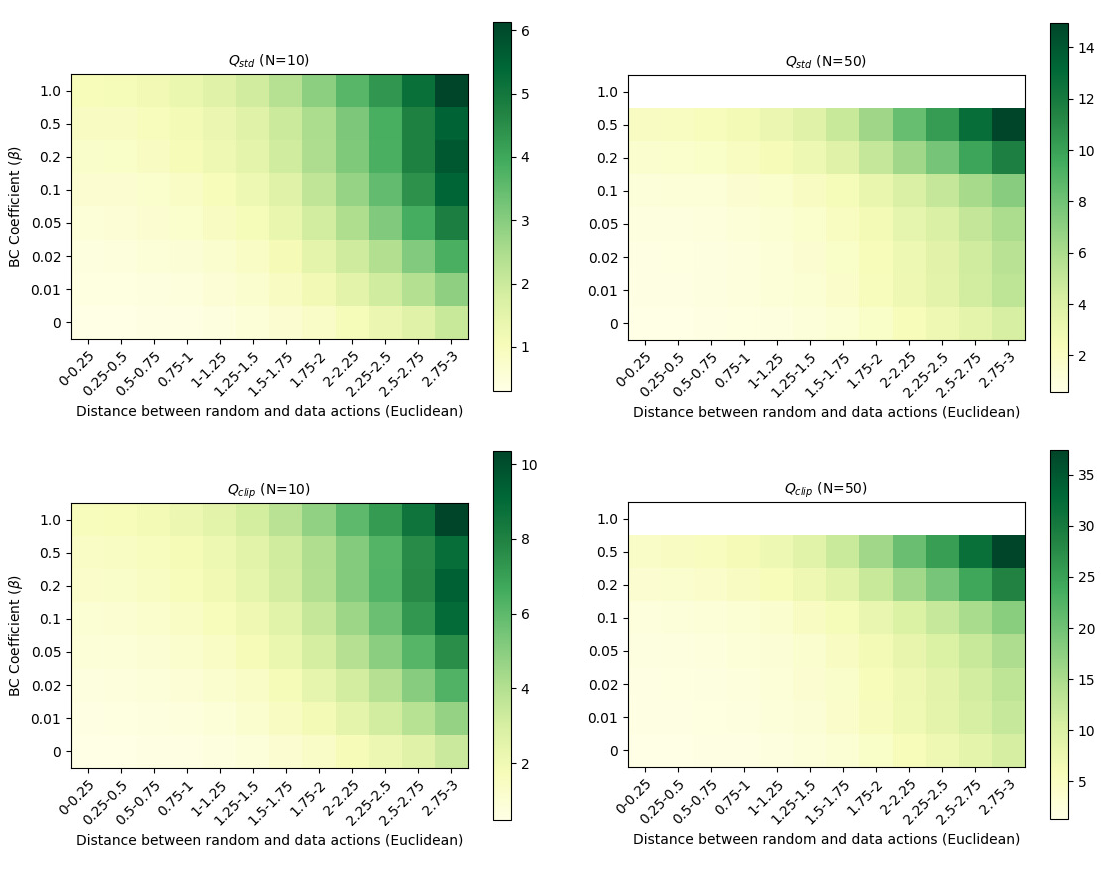}
            \end{center}
            \caption{Uncertainty as a function of distance, $N$ and $\beta$. Top row standard deviation, bottom row clip penalty.  As the distance between random and data actions increases so too does the level of uncertainty, becoming more pronounced as $\beta$ and $N$ get larger.  White space is used to represent erroneous values due to unreliable Q-values estimates resulting from divergent critic loss during training}
            \label{fig:StdClipPen}
        \end{figure}

    To this effect, we sample 50000 states from the data and 50000 actions from a random policy and calculate (a) the Euclidean distance between random and data actions and (b) the standard deviation/clip penalty.  We then group distances into equally sized bins and within each bin calculate the average standard deviation/clip penalty.  We summarise results for $N=10$ and $N=50$ in Figure \ref{fig:StdClipPen} via heatmaps, using shade to represent the size of the corresponding uncertainty metric.  Similar plots for $N=[2, 5, 20]$ can be found in the Appendix.  In general, we see that as the distance between random and data actions increases, so too does the level of uncertainty (standard deviation and penalty gap), and this becomes more pronounced as the value of $\beta$ increases.  This supports our hypothesis that policy constraints can be used to control uncertainty in Q-value estimates.  We also see that the highest levels of uncertainty occur when both $N$ and $\beta$ are large, supporting our explanation of declining performance as observed in Figure \ref{fig:EnsVsBeta}.

\begin{figure}[h] % h=here, t=top, b=bottom, p=new page, !=other preferences
            \begin{center}
                \includegraphics[width=1\textwidth]{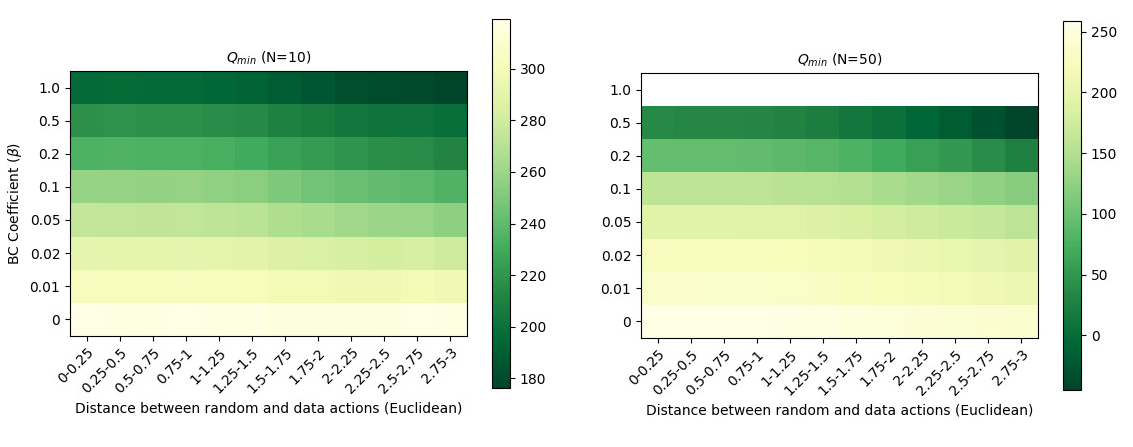}
            \end{center}
            \caption{$Q_{min}$ as a function of distance, $N$ and $\beta$  As the distance between random and data actions increases, $Q_{min}$ decreases, with this decrease more pronounced as $\beta$ and $N$ get larger.  White space is used to represent erroneous values due to unreliable Q-values estimates resulting from divergent critic loss during training}
            \label{fig:HeatMin}
        \end{figure}

    Finally, we also examine the distribution of the minimum across the ensemble, $Q_{min}$, as this value is the one used in updates during policy evaluation and policy improvement.  Using the same format as for uncertainty, we summarise results for $N=10$ and $N=50$ in Figure \ref{fig:HeatMin}, using shade to represent the value of $Q_{min}$.  In general, we see that $Q_{min}$ decreases as the distance between random and data actions increases, being more pronounced as either $N$ or $\beta$ increase.  This culminates in the lowest $Q_{min}$ values when the size of the ensemble and level of constraint are at their highest, mirroring the findings based on uncertainty measures.

    For completeness, we reproduce these plots for agents trained using independent target values in the Appendix, finding in general the same features.  We also provide additional plots examining (a) the distribution of $Q_{min}$ for policy actions and (b) the shape of the distribution of Q-value estimates for individual actions, providing more insights into the impact of $\beta$ on uncertainty.
   
     \subsection{Performance and efficiency comparisons} \label{performance}
    As one of our objectives is to attain the same-level of performance as ensemble-based methods, we compare to published results from SAC-N, EDAC and MSG.  As the leading BC based approaches we also compare to published results from IQL and TD3-BC.  Finally, since MSG makes use of CQL we also compare to updated results as published in the IQL paper\footnote{These results are based on updated D4RL data sets following minor bug fixes}.
    
\begin{figure}[] % h=here, t=top, b=bottom, p=new page, !=other preferences
            \begin{center}
                \includegraphics[width=1\textwidth]{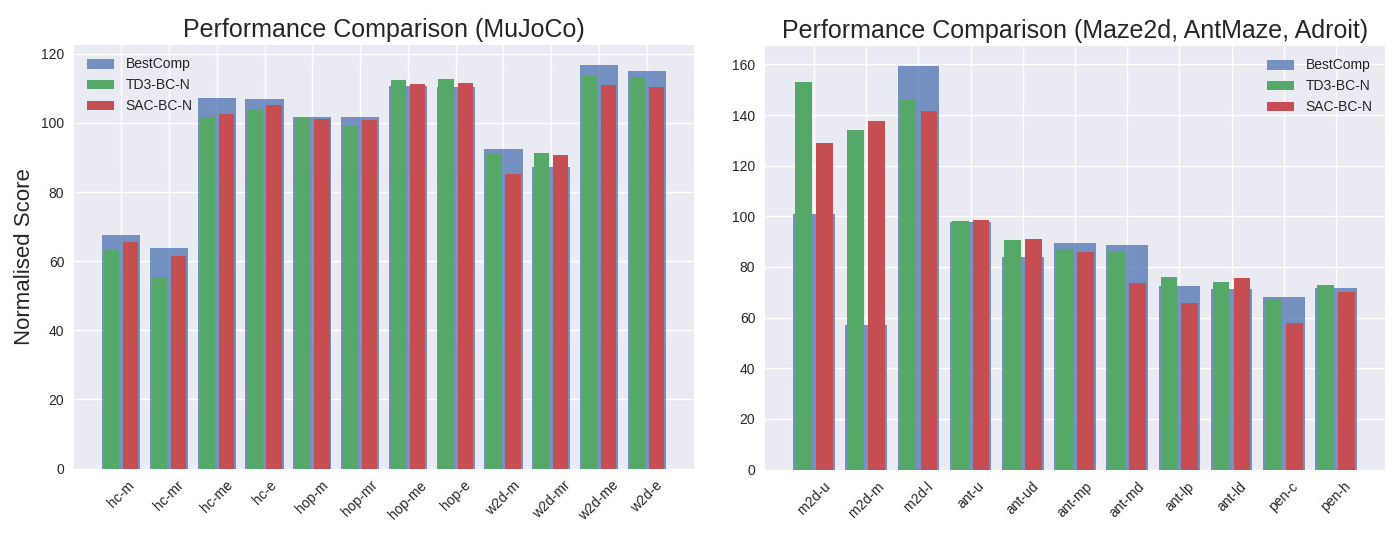}
            \end{center}
            \caption{Comparing the performance of TD3-BC-N (green) and SAC-BC-N (red) against the best method from Table 1 (blue).  Performance is competitive across all tasks}
            \label{fig:offline_results}
        \end{figure}

    We present results for all tasks and data sets in Table \ref{tbl_offline_results}.  Where figures are not published for a given task we denote the entry as ``-''\footnote{While MSG does consider the MuJoCo environments, results are only presented visually and are in general on-par or below those of SAC-N/EDAC.}.  To help better visualise performance levels,  we compare our results to the best performing method in Figure \ref{fig:offline_results}, which with a few exceptions is SAC-N/EDAC for MuJoCo and Adroit, and MSG for maze tasks.  For the MuJoCo and Adroit environments, we see that in general both TD3-BC-N and SAC-BC-N can match the performance of SAC-N and EDAC, and for Maze2d and AntMaze they can match the performance of MSG.  Note that this is achieved without adjusting hyperparameters within the same task, in contrast to SAC-N, EDAC and MSG.  In the Appendix we investigate the effect of removing this restriction using the MuJoCo environments, finding performance can be slightly enhanced.
    
\begin{figure}[] % h=here, t=top, b=bottom, p=new page, !=other preferences
            \begin{center}
                \includegraphics[width=1\textwidth]{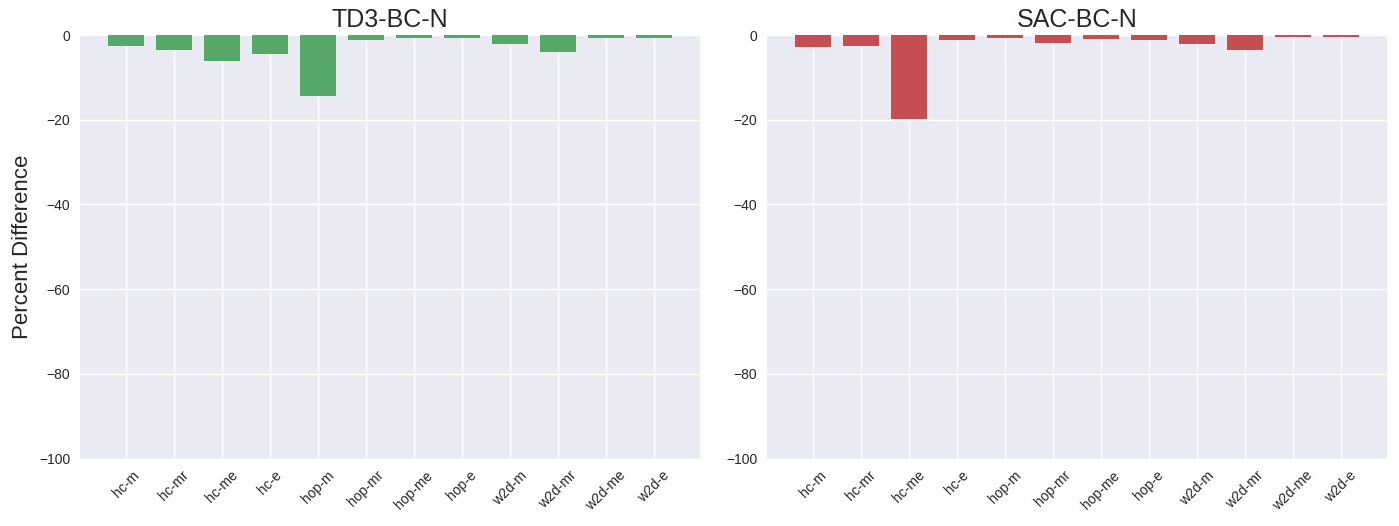}
            \end{center}
            \caption{Evaluating robustness of learn policies for MuJoCo tasks.  Each plot shows the percentage difference between the mean and worst performing episode across 50 evaluations (10 evaluations per 5 seeds).  With the exception of one data set, both TD3-BC-N and SAC-BC-N are able to produce robust policies regardless of data quality}
            \label{fig:offline_results_worst}
        \end{figure}

    For the MuJoCo domain in particular we note there is very little variation in performance across seeds/evaluations, demonstrating our approach is able to learn robust as well as performant policies.  This is further evidenced in Figure \ref{fig:offline_results_worst} where we plot the percentage difference between the mean and worst score across the 50 evaluations, which in most cases is negligible.  Since real-world application will typically only involve single policy deployment, such a property is highly desirable.

    \begin{table}[]
    \begin{adjustbox}{width=\textwidth}
    \begin{tabular}{|l||cccccc|cc|}
    \hline
    \textbf{Task/data set} & CQL   & IQL   & \begin{tabular}[c]{@{}c@{}}TD3\\ -BC\end{tabular} & EDAC  & \begin{tabular}[c]{@{}c@{}}SAC\\ -N\end{tabular} & MSG   & \begin{tabular}[c]{@{}c@{}}TD3\\ -BC-N\end{tabular} & \begin{tabular}[c]{@{}c@{}}SAC\\ -BC-N\end{tabular} \\ \hline
    halfcheetah-v2       &       &       &                                                   &       &                                                  &       &                                                     &                                                     \\
    -medium           & 44.0  & 47.4  & 48.3                                              & 65.9  & 67.5                                             & -     & 63.3 $\pm{0.1}$                                     & 65.6 $\pm{0.2}$                                     \\
    -medium-replay    & 45.5  & 44.2  & 44.6                                              & 61.3  & 63.9                                             & -     & 55.3 $\pm{0.1}$                                     & 61.5 $\pm{0.1}$                                     \\
    -medium-expert    & 91.6  & 86.7  & 90.7                                              & 106.3 & 107.1                                            & -     & 101.7 $\pm{0.3}$                                    & 102.6 $\pm{0.5}$                                    \\
    -expert           & -     & -     & 96.7                                              & 106.8 & 105.2                                            & -     & 103.8 $\pm{0.5}$                                    & 105.3 $\pm{0.1}$                                    \\ \hline
    hopper-v2            &       &       &                                                   &       &                                                  &       &                                                     &                                                     \\
    -medium           & 58.5  & 66.3  & 59.3                                              & 101.6 & 100.3                                            & -     & 101.5 $\pm{0.3}$                                    & 101.2 $\pm{0.1}$                                    \\
    -medium-replay    & 95.0  & 94.7  & 60.9                                              & 101.0 & 101.8                                            & -     & 99.1 $\pm{0.1}$                                     & 100.8 $\pm{0.2}$                                    \\
    -medium-expert    & 105.4 & 91.5  & 98.0                                              & 110.7 & 110.1                                            & -     & 112.3 $\pm{0.0}$                                    & 111.3 $\pm{0.1}$                                    \\
    -expert           & -     & -     & 107.8                                             & 110.1 & 110.3                                            & -     & 112.7 $\pm{0.1}$                                    & 111.5 $\pm{0.1}$                                    \\ \hline
    walker2d-v2          &       &       &                                                   &       &                                                  &       &                                                     &                                                     \\
    -medium           & 72.5  & 78.3  & 83.7                                              & 92.5  & 87.9                                             & -     & 90.9 $\pm{0.2}$                                     & 85.3  $\pm{0.1}$                                    \\
    -medium-replay    & 77.2  & 73.9  & 81.8                                              & 87.1  & 78.7                                             & -     & 91.4 $\pm{0.4}$                                     & 90.8 $\pm{0.2}$                                     \\
    -medium-expert    & 108.8 & 109.6 & 110.1                                             & 114.7 & 116.7                                            & -     & 113.5 $\pm{0.1}$                                    & 110.9 $\pm{0.0}$                                    \\
    -expert           & -     & -     & 110.2                                             & 115.1 & 107.4                                            & -     & 113.2 $\pm{0.0}$                                    & 110.4 $\pm{0.0}$                                    \\ \hline
    mujoco average    & -     & -     & 82.3                                              & 97.8  & 96.4                                             & -     & 96.6                                                & 96.4                                                \\
    (exc. expert)     & 77.6  & 77.0  & 75.3                                              & 93.5  & 92.7                                             & -     & 92.2                                                & 92.2                                                \\ \hline \hline
    maze2d-v1            &       &       &                                                   &       &                                                  &       &                                                     &                                                     \\
    -umaze            & -     & -     & -                                                 & -     & -                                                & 101.1 & 153.1 $\pm{1.5}$                                    &   128.9 $\pm{1.7}$                                                  \\
    -medium           & -     & -     & -                                                 & -     & -                                                & 57.0  & 133.9 $\pm{2.0}$                                     & 137.8 $\pm{2.1}$                                                    \\
    -large            & -     & -     & -                                                 & -     & -                                                & 159.3 & 145.8 $\pm{3.3}$                                    &  141.4 $\pm{3.6}$                                                   \\ \hline
    maze2d average    & -     & -     & -                                                 & -     & -                                                & 105.8 & 144.3                                               &  136.0                                                   \\ \hline \hline
    antmaze-v0           &       &       &                                                   &       &                                                  &       &                                                     &                                                     \\
    -umaze            & 74.0  & 87.5  & 78.6                                              & -     & -                                                & 97.8  & 98.3 $\pm{0.7}$                                     &            98.6 $\pm{0.5}$                                         \\
    -umaze-diverse    & 84.0  & 62.2  & 71.4                                              & -     & -                                                & 81.8  & 90.6 $\pm{1.3}$                                     &            91.2 $\pm{1.3}$                                         \\
    -medium-play      & 61.2  & 71.2  & 10.6                                              & -     & -                                                & 89.6  & 87.0 $\pm{1.5}$                                     &            85.8 $\pm{1.6}$                                         \\
    -medium diverse   & 53.7  & 70.0  & 3.0                                               & -     & -                                                & 88.6  & 86.2 $\pm{1.5}$                                     &            73.8 $\pm{2.0}$                                         \\
    -large-play       & 15.8  & 39.6  & 0.2                                               & -     & -                                                & 72.6  & 76.2 $\pm{1.9}$                                     &            65.8 $\pm{2.1}$                                         \\
    -large-diverse    & 14.9  & 47.5  & 0                                                 & -     & -                                                & 71.4  & 74.2 $\pm{2.0}$                                     &            75.8 $\pm{1.9}$                                         \\ \hline
    antmaze average   & 50.6  & 63    & 27.3                                              & -     & -                                                & 83.6  & 85.5                                                &            81.8                                         \\ \hline \hline
    adroit-v1            &       &       &                                                   &       &                                                  &       &                                                     &                                                     \\
    -pen-cloned       & 39.2  & 37.3  & -                                                 & 68.2  & 64.1                                             & -     & 67.2 $\pm{2.9}$                                     &        58.0 $\pm{2.8}$                                             \\
    -pen-human        & 37.5  & 71.5  & -                                                 & 52.1  & 9.5                                              & -     & 72.8 $\pm{2.7}$                                     &        70.3 $\pm{2.9}$                                             \\ \hline
    adroit average    & 38.4  & 54.4  & -                                                 & 60.2  & 36.8                                             & -     & 70.0                                               &        64.2                                             \\ \hline
    \end{tabular}
    \end{adjustbox}
    \caption{Performance comparison across D4RL benchmark.  Figures are normalised scores, with 0 and 100 representing random and expert policies, respectively. 
 For TD3-BC-N and SAC-BC-N we report the mean normalised score $\pm$ one standard error across 50 evaluations for MuJoCo tasks (10 evaluations over 5 seeds) and 500 evaluations for Maze2d, AntMaze and Adroit tasks (100 evaluations over 5 seeds). 
 Both TD3-BC-N and SAC-BC-N are able to match the state-of-the-art performance across all domains.  This is the case even with the restriction preventing BC adjustments within the same task}\label{tbl_offline_results}
    \end{table}

    After demonstrating our approach can match state-of-the-art alternatives in terms of performance, we turn our attention to computational efficiency.  To ensure a fair comparison, we implement our own versions of baselines based on author published source code and the CORL repository \citep{tarasov2022corl}, and run them on the same hardware/software configuration.  We use exactly the same network architecture across ensemble-based approaches, training each member of the ensemble in parallel.  For CQL, IQL and TD3-BC we use the network architecture as described in their respective papers.  Full details are provided in the Appendix.
    
    In Figure \ref{fig:comp_eff} we plot the training time in hours of each approach, considering several variations of SAC-N, EDAC and MSG based on ensemble size, which varies according to task type.  We see that TD3-BC-N and SAC-BC-N are easily the most efficient among the ensemble-based approaches, a direct consequence of a smaller ensemble size and need for fewer gradient updates to reach peak performance.  In particular, the computation time for TD3-BC-N is comparable to the minimalist approach of TD3-BC.
    
    \begin{figure}[h] % h=here, t=top, b=bottom, p=new page, !=other preferences
            \begin{center}
                \includegraphics[width=1\textwidth]{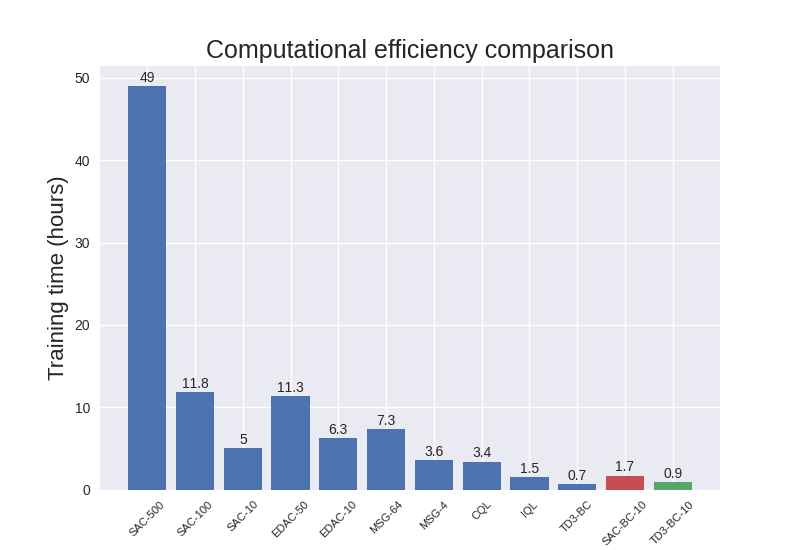}
            \end{center}
            \caption{Computational efficiency.  A smaller ensemble size coupled with fewer gradient updates allows TD3-BC-N and SAC-BC-N to significantly reduce computation time to levels similar to that of more minimalist approaches such as TD3-BC}
            \label{fig:comp_eff}
        \end{figure}

\begin{figure}[t] % h=here, t=top, b=bottom, p=new page, !=other preferences
            \begin{center}
                \includegraphics[width=1\textwidth]{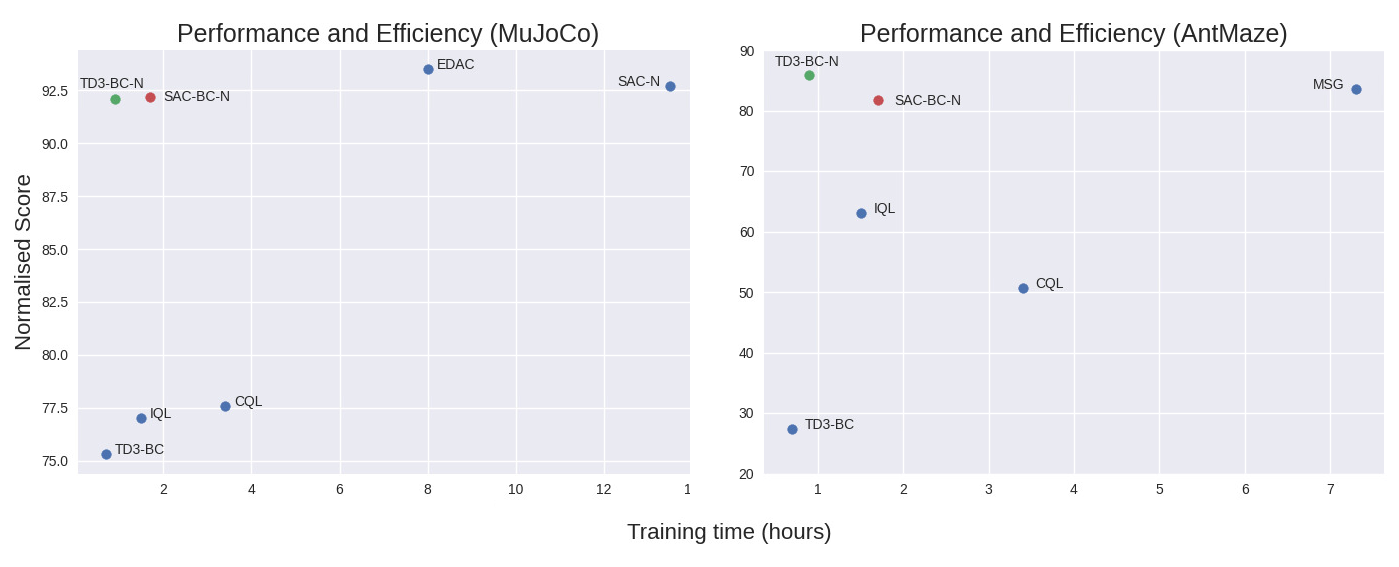}
            \end{center}
            \caption{Performance and efficiency.  Average training time and normalised score across MuJoCo and AntMaze tasks.  TD3-BC-N and SAC-BC-N can match the performance of ensemble-based approaches while retaining the computational efficiency of those based on behavioural cloning}
            \label{fig:TTvsNS}
        \end{figure}

    To get a clearer sense of how performance and efficiency compare across algorithms, in Figure \ref{fig:TTvsNS} we plot the average training time and normalised score for MuJoCo\footnote{We exclude the ``expert'' data sets since these are not reported for CQL or IQL} and AntMaze tasks.  We see that ensemble-based approaches (SAC-N, EDAC, MSG) are the most performant, but also the most computationally expensive.  Conversely, BC based approaches (TD3-BC, IQL) are the most computationally efficient, but least performant.  TD3-BC-N and SAC-BC-N on other hand are able to retain the advantages of both approaches while diminishing their individual deficiencies.

    \subsubsection*{Ablation studies}
    In addition to our main results, we also conduct a number of ablations studies to verify the importance of individual components of our approach, as well implementation decisions.  In Ablations 1-3, we use the MuJoCo environments to assess the impact of removing the BC component, ensemble of critics and inflated period of BC, respectively.  In Ablation 4, we use the AntMaze and Adroit environments to show the impact of using dependent targets instead of independent targets during policy evaluation.  We conduct these ablations using TD3-BC-N, making no other changes than those outlined above.

\begin{figure}[] % h=here, t=top, b=bottom, p=new page, !=other preferences
            \begin{center}
                \includegraphics[width=1\textwidth]{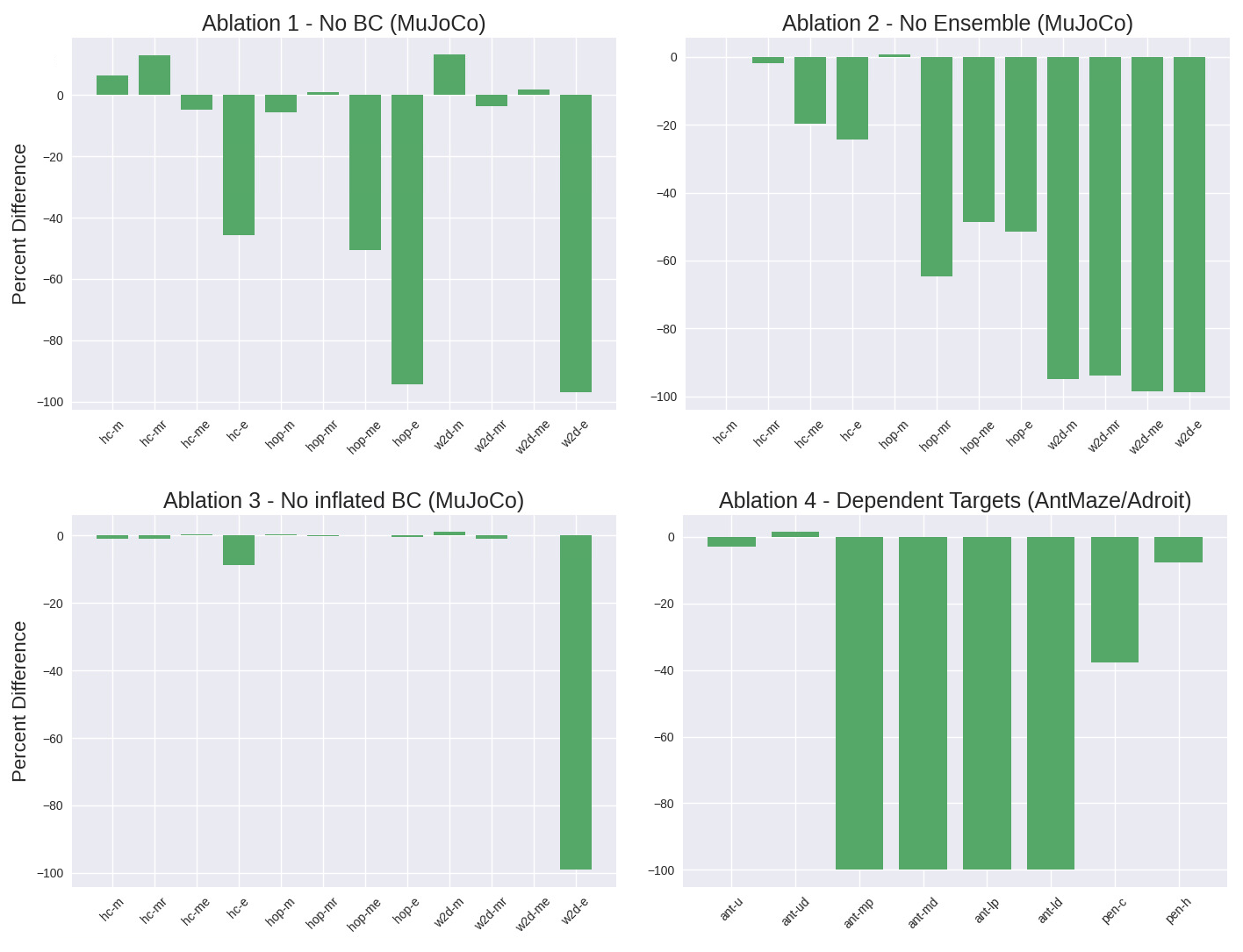}
            \end{center}
            \caption{Ablations studies.  Each plot shows the percentage difference in mean normalised score between each ablation and the main results from Table 1.  Ablations 1 and 2 show that both behavioural cloning and an ensemble of critics are necessary to achieve strong performance.  Ablations 3 and 4 show the importance of our implementation choices, namely the use of an initial period of inflated BC and independent targets for AntMaze/Adroit environments}
            \label{fig:ablations}
        \end{figure}

    We summarise results in Figure \ref{fig:ablations}, plotting the percentage difference between each ablation score and the main results of Table \ref{tbl_offline_results}.  For Ablations 1-2 we see that removing either BC or the ensemble has a detrimental impact on performance overall.  While the performance for some tasks is unaffected by removing the BC component, there are others that suffer catastrophic failure and hence its inclusion is essential.  For Ablation 3 we see removing the inflated period of BC has minimal impact on most data sources, but the severe impact on ``walker2d-expert'' warrants its inclusion.  Finally, in Ablation 4 we see the use of independent targets is crucial for the more challenging ``medium'' and ``large'' AntMaze environments and is beneficial for Adroit environments.

     \subsection{Online fine-tuning}\label{Expts_OnlineFT}   
    Starting with our offline trained agents, we perform online fine-tuning according to the procedures outlined in Algorithms \ref{algoTD3-BC-N-FT} and \ref{algoSAC-BC-N-FT}.  We populate the replay buffer $R$ with the last 2500 transitions from $D$ and train agents for an additional 250k environment interactions, with gradient updates commencing after the first 2500 interactions (i.e. $K=2500$).  With the exception of the ``maze2d-umaze'' environment where $\beta_{start}=0.2$, the offline value of $\beta$ is used for $\beta_{start}$ and the number of decay steps $S$ is set as 50k.  The value of $\beta_{end}$ is set according to environment and procedure, but as with our offline experiments, its value doesn't change according to initial data quality.  Values for each data set and procedure are provided in the Appendix.  All other parameters remain the same.

\begin{figure}[] % h=here, t=top, b=bottom, p=new page, !=other preferences
            \begin{center}
                \includegraphics[width=1\textwidth]{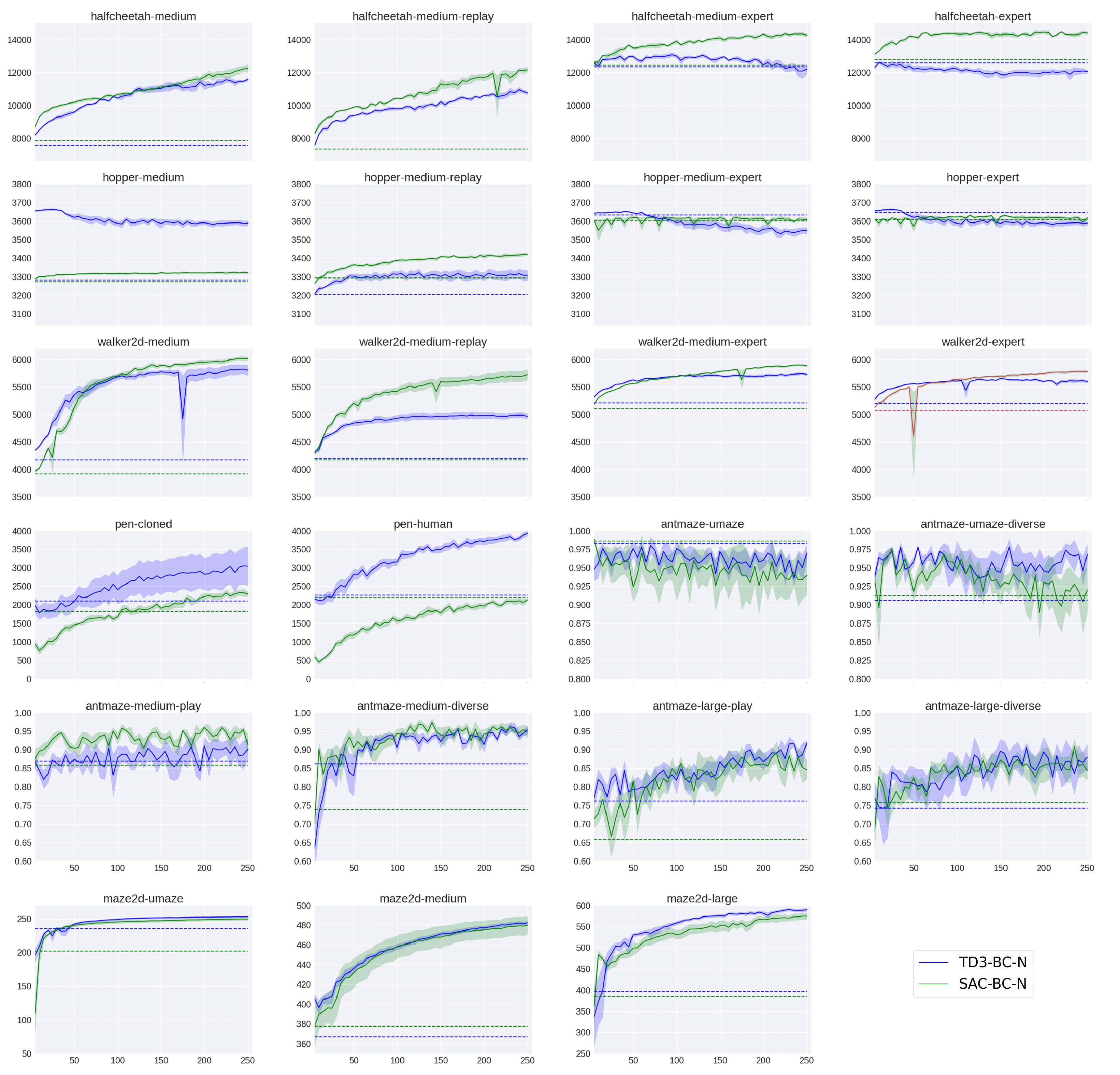}
            \end{center}
            \caption{Online fine-tuning for D4RL tasks.  The solid line represents the mean non-normalised score across each of the five agents, shaded area the standard error and dashed line performance prior to fine-tuning.  In general, agents are able to improve their policies in a stable manner, with only a few tasks/data sources causing stability issues}
            \label{fig:OnlineFT}
        \end{figure}

    For each task, we plot the corresponding learning curves in Figure \ref{fig:OnlineFT}, evaluating policies every 5000 environment interactions (10 evaluations for MuJoCo, 100 evaluations for Adroit/AntMaze/Maze2d).  The solid line represents the mean (non-normalised) score across each of the five seeds, shaded area the standard error and dashed line performance prior to fine-tuning.  For the MuJoCo environment, in the majority of cases agents are able to improve their policies while avoiding severe performance drops during the offline to online transition.  For TD3-BC-N, the performance for ``hopper/halfcheetah-expert'' declines slightly over the course of training and for SAC-BC-N there is sharp decline for ``walker2d-expert'' within the $\beta$ decay period.  For Adroit, TD3-BC-N manages a reasonable transition and subsequent improvement, but SAC-BC-N is less successful, particularly for ``pen-cloned''.  With the exception of ``antmaze-umaze'', in AntMaze both TD3-BC-N and SAC-BC-N obtain improved policies in a reasonable stable manner.  Finally, for Maze2d we see continued improvement for both methods, with some minor initial deterioration in TD3-BC-N for ``maze2d-umaze'' and ``maze2d-large'' and fairly large initial slump in SAC-BC-N for ``maze2d-umaze''
    
\section{Discussion and conclusion}\label{DiscConc}
In this work we have investigated the role of policy constraints as a mechanism for improving the computational efficiency of ensemble-based approached to offline reinforcement learning.  Through empirical evaluation, we have shown how constraints in the form of behavioural cloning can be used to control the level of uncertainty in the estimated value of out-of-distribution actions, allowing these estimates to be sufficiently penalised to prevent overestimation bias.  Through this feature, we have been able to match state-of-the-art performance across a number of challenging benchmarks while significantly reducing computational burden, cutting the size of the ensemble to a fraction of that needed when policies are unconstrained.  We have also shown how behavioural cloning can be repurposed to promote stable and performant online fine-tuning, by gradually reducing its influence during the offline-to-online transition.  These achievements have required only minimal changes to existing approaches, allowing for easy implementation and interpretation.

Our work highlights a number of interesting avenues for future research.  Primary among these is the development of methods for selecting the size of the ensemble $N$ and level of behavioural cloning $\beta$ offline.  While we have demonstrated our approach can achieve strong performance using consistent hyperparameters, we have also shown how performance can be further improved by allowing them to vary.  Related to this is the development of approaches for automatically tuning $\beta$ during training, possibly making use of uncertainty metrics described in Section \ref{interpretation}.  A theoretical analysis of the impact of $\beta$ on uncertainty could also prove beneficial in this regard.  

While in this work we have used ensembles for uncertainty estimation, other techniques such a multi-head, multi-input/outputs and Monte Carlo dropout can just as easily be used and integrated with BC.  Similarly, other forms of policy constraints and/or other divergence metrics can be incorporated into ensemble-based approaches in a relatively straightforward manner.  As such, there a number of permutations which could lead to improved performance and/or computational efficiency.  

Finally, our fine-tuning procedure may benefit from incorporating elements from methods outlined in Section \ref{RelatedWork}, allowing for greater stability during the entire duration of online learning.  In addition, our approach may also prove useful in promoting greater data efficiency in online-RL.

\backmatter

% \bmhead{Supplementary information}

% Put GitHub links here (already in footnote)

\bmhead{Acknowledgments}
\noindent
AB acknowledges support from University of Warwick and University of Birmingham NHS Foundation Trust. 
\noindent
GM acknowledges support from a UKRI AI Turing Acceleration Fellowship (EPSRC EP/V024868/1). 
\noindent
The authors acknowledge Weights \& Biases (https://www.wandb.com/) as the online platform used for experiment tracking and visualizations to develop insights for this paper.

\bibliography{refs}% common bib file

\newpage

\section*{Statements and Declarations}

\subsection*{Funding}
AB acknowledges support from University of Warwick and University of Birmingham NHS Foundation Trust. 
GM acknowledges support from a UKRI AI Turing Acceleration Fellowship (EPSRC EP/V024868/1). 

\subsection*{Conflict of interest/Competing interests}
No competing or financial interests to disclose.

\subsection*{Ethics approval}
Not applicable.

\subsection*{Consent to participate}
The authors give their consent to participate.

\subsection*{Consent for publication}
The authors give their consent for publication.

\subsection*{Availability of data and materials}
Benchmark data sets are open source.  Code base for implementation is made freely available.

\subsection*{Authors' contributions}
Authors' contributions follow the authors' order convention.

\newpage

\section*{Appendix}
\subsection*{SAC-BC-N online fine-tuning procedure}\label{AppOnlineFT}
Following on from Section \ref{OnlineFT}, we outline the online fine-tuning procedure using SAC-BC-N in Algorithm \ref{algoSAC-BC-N-FT}.

\begin{algorithm}[]
       \caption{Online fine-tuning (SAC-BC-N)}\label{algoSAC-BC-N-FT}
    \begin{algorithmic}
       \State {\bfseries Require:} Ensemble size $N$, discount factor $\gamma$, minimum entropy $\mathcal{H}$, target network update rate $\tau$, data set $D$ and decay parameters $\beta_{start}, \beta_{end}, S$
       \State Initialise pre-trained critic parameters $\theta_i$, policy parameters $\phi$ and corresponding target parameters $\theta'_i$.
       \State Initialise environment and replay buffer $R$
       \State Populate $R$ with a proportion of transitions from $D$.
        \For {$k=0$ to $K$}
           \State Act in environment with exploration, $a \sim \pi_\phi(s)$
           \State Store resulting transition $(s, a, r, s')$ in $R$
        \EndFor
       \State Set decay rate $\kappa_{\beta}$ as per equation (\ref{beta_decay})
       \State Set $\beta = \beta_{start}$
       \For {$j=0$ to $J$}
           \State Act in environment with exploration, $a \sim \pi_\phi(s)$
           \State Store resulting transition $(s, a, r, s')$ in $R$
           \State Sample minibatch of transitions $(s, a, r, s')$ from $R$
            \State Update Q-function parameters $\theta_i$ using equation (\ref{SAC-BC-N-PolEvalDep}) or (\ref{SAC-BC-N-PolEvalIndep})
            \State Update policy parameters $\phi$ using equation (\ref{SAC-BC-N-PolImpMSE}) or (\ref{SAC-BC-N-PolImpLL})
            \State Update entropy parameter $\alpha$ using equation (\ref{SAC-BC-N-Ent})
       \State Update target network parameters $\theta'_i$ using equation (\ref{SAC-BC-N-Targets})
           \State Update BC coefficient $\beta = \max(\beta, \kappa_{\beta}\beta$)
       \EndFor
    \end{algorithmic}
    \label{alg:td3-bc-n}
    \end{algorithm}

\subsection*{Further implementation details}\label{AppImp}
As per previous works, we perform the following data transformations:
\begin{itemize}
    \item Normalise states as per TD3-BC
    \item Transform AntMaze rewards according to $4(r - 0.5)$ as per MSG/CQL
    \item Normalise Adroit rewards as per SAC-N/EDAC
\end{itemize}

\subsubsection*{TD3-BC-N hyperparameters and network architecture}  
Following on from Section \ref{Expts}, we provide details of shared hyperparameters and network architecture in Table \ref{TD3-BC-N-Gen}, and details of task specific hyperparameters for BC in Table \ref{TD3-BC-N-Spec}.

\begin{table}[]
\begin{center}
\begin{tabular}{cll}
\hline
\multicolumn{1}{l}{}           & \textbf{Hyperparameter}      & \textbf{Value}            \\ \hline
\multirow{9}{*}{TD3-BC-N}       & Optimiser                    & Adam                      \\
                               & Actor learning rate          & 3e-4                      \\
                               & Critic learning rate         & 3e-4                      \\
                               & Batch size                   & 256                       \\
                               & Discount factor $\gamma$             & 0.99                      \\
                               & Target network update rate $\tau$   & 0.005                     \\
                               & Policy noise $\epsilon$                 & 0.2                       \\
                               & Policy noise clipping        & (-0.5, 0.5)               \\
                               & Critic-to-Actor update ratio & 2:1                       \\ \hline
\multirow{2}{*}{TD3-BC-N online} & Exploration noise $\sigma$ & 0.1                     \\
                               & BC decay stay steps $S$ & 50,000                      \\ \hline
\multirow{11}{*}{Architecture} & Critic hidden nodes          & 256                       \\
                               & Critic hidden layers         & 3                         \\
                               & Critic hidden activation     & ReLU                      \\
                               & Critic input                 & State + Action            \\
                               & Critic output                & Q-value                   \\
                               & Ensemble size $N$            & 10                        \\
                               & Actor hidden nodes           & 256                       \\
                               & Actor hidden layers          & 3                         \\
                               & Actor hidden activation      & ReLU                      \\
                               & Actor input                  & State                     \\
                               & Actor outputs                & Action (tanh transformed) \\ \hline
\end{tabular}
\caption{TD3-BC-N shared hyperparameters and network architecture}
\label{TD3-BC-N-Gen}
\end{center}
\end{table}

\begin{table}[]
\begin{center}
\begin{tabular}{|ccll|}
\hline
\multicolumn{1}{|c}{\textbf{Task}} & \textbf{Dataset} & \textbf{$\beta$} & \textbf{$\beta_{end}$} \\ \hline
\multirow{4}{*}{halfcheetah}       & medium           & 0.04             & $1e^{-12}$                 \\
                                   & medium-replay    & 0.04             & $1e^{-12}$                 \\
                                   & medium-expert    & 0.04             & $1e^{-12}$                \\
                                   & expert           & 0.04             & $1e^{-12}$               \\ \hline
\multirow{4}{*}{hopper}            & medium           & 0.03             & 0.02                   \\
                                   & medium-replay    & 0.03             & 0.02                   \\
                                   & medium-expert    & 0.03             & 0.02                   \\
                                   & expert           & 0.03             & 0.02                   \\ \hline
\multirow{4}{*}{walker2d}          & medium           & 0.03             & $1e^{-10}$               \\
                                   & medium-replay    & 0.03             & $1e^{-10}$               \\
                                   & medium-expert    & 0.03             & $1e^{-10}$               \\
                                   & expert           & 0.03             & $1e^{-10}$               \\ \hline
maze2d-umaze                       &                  & 0.02             & 0.02                   \\ \hline
maze2d-medium                      &                  & 0.2              & 0.02                   \\ \hline
maze2d-large                       &                  & 0.2              & 0.02                   \\ \hline
\multirow{2}{*}{antmaze-umaze}     & -                & 0.1              & 0.1                    \\
                                   & -diverse         & 0.1              & 0.1                    \\ \hline
\multirow{2}{*}{antmaze-medium}    & -play            & 0.02             & 0.01                   \\
                                   & -diverse         & 0.02             & 0.01                   \\ \hline
\multirow{2}{*}{antmaze-large}     & -play            & 0.02             & 0.005                  \\
                                   & -diverse         & 0.02             & 0.005                  \\ \hline
\multirow{2}{*}{pen}               & -cloned          & 10               & 2                      \\
                                   & -human           & 10               & 2                      \\ \hline
\end{tabular}
\caption{TD3-BC-N task specific BC hyperparameters.  Note the BC parameters are fixed within each task, i.e. do not vary based on dataset}
\label{TD3-BC-N-Spec}
\end{center}
\end{table} 

\subsubsection*{SAC-BC-N hyperparameters and network architecture}  
Following on from Section \ref{Expts}, we provide details of shared hyperparameters and network architecture in Table \ref{SAC-BC-N-Gen}, and details of task specific hyperparameters for BC in Table \ref{SAC-BC-N-Spec}.

\begin{table}[]
\begin{center}
\begin{tabular}{cll}
\hline
\multicolumn{1}{l}{}           & \textbf{Hyperparameter}      & \textbf{Value}            \\ \hline
\multirow{7}{*}{SAC-BC-N}       & Optimiser                    & Adam                      \\
                               & Actor learning rate          & 3e-4                      \\
                               & Critic learning rate         & 3e-4                      \\
                               & Batch size                   & 256                       \\
                               & Discount factor $\gamma$             & 0.99                      \\
                               & Target network update rate $\tau$   & 0.005                     \\
                               & Minimum entropy $H$                 & -1 * action dimension                       \\ \hline
\multirow{1}{*}{SAC-BC-N online} & BC decay stay steps $S$ & 50,000                     \\ \hline
\multirow{11}{*}{Architecture} & Critic hidden nodes          & 256                       \\
                               & Critic hidden layers         & 3                         \\
                               & Critic hidden activation     & ReLU                      \\
                               & Critic input                 & State + Action            \\
                               & Critic output                & Q-value                   \\
                               & Ensemble size $N$            & 10                        \\
                               & Actor hidden nodes           & 256                       \\
                               & Actor hidden layers          & 3                         \\
                               & Actor hidden activation      & ReLU                      \\
                               & Actor input                  & State                     \\
                               & Actor outputs                & Mean/standard deviation of Gaussian \\ \hline
\end{tabular}
\caption{SAC-BC-N shared hyperparameters and network architecture}
\label{SAC-BC-N-Gen}
\end{center}
\end{table}

% \usepackage{multirow}
\begin{table}[]
\begin{center}
\begin{tabular}{|cccll|}
\hline
\multicolumn{1}{|c}{\textbf{Task}} & \textbf{Dataset} & \textbf{BC form} & \textbf{$\beta$} & \textbf{$\beta_{end}$} \\ \hline
\multirow{4}{*}{halfcheetah}       & medium           & Log-likelihood   & 0                & 0                      \\
                                   & medium-replay    & Log-likelihood   & 0                & 0                      \\
                                   & medium-expert    & Log-likelihood   & 0                & 0                      \\
                                   & expert           & Log-likelihood   & 0                & 0                      \\ \hline
\multirow{4}{*}{hopper}            & medium           & Log-likelihood   & 0.0025           & 0.001                  \\
                                   & medium-replay    & Log-likelihood   & 0.0025           & 0.001                  \\
                                   & medium-expert    & Log-likelihood   & 0.0025           & 0.001                  \\
                                   & expert           & Log-likelihood   & 0.0025           & 0.001                  \\ \hline
\multirow{4}{*}{walker2d}          & medium           & Log-likelihood   & 0.0025           & $1e^{-10}$               \\
                                   & medium-replay    & Log-likelihood   & 0.0025           & $1e^{-10}$               \\
                                   & medium-expert    & Log-likelihood   & 0.0025           & $1e^{-10}$               \\
                                   & expert           & Log-likelihood   & 0.0025           & $1e^{-10}$               \\ \hline
maze2d-umaze                       &                  & MSE              & 0.02             & 0.02                       \\ \hline
maze2d-medium                      &                  & MSE              & 0.05             & 0.01                       \\ \hline
maze2d-large                       &                  & MSE              & 0.05             & 0.01                       \\ \hline
\multirow{2}{*}{antmaze-umaze}     & -                & MSE              & 0.1              & 0.05                       \\
                                   & -diverse         & MSE              & 0.1              & 0.05                       \\ \hline
\multirow{2}{*}{antmaze-medium}    & -play            & MSE              & 0.02             & 0.02                       \\
                                   & -diverse         & MSE              & 0.02             & 0.02                       \\ \hline
\multirow{2}{*}{antmaze-large}     & -play            & MSE              & 0.01             & 0.005                       \\
                                   & -diverse         & MSE              & 0.01             & 0.005                       \\ \hline
\multirow{2}{*}{pen}               & -cloned          & MSE              & 10                & 2                       \\
                                   & -human           & MSE              & 10                & 2                       \\ \hline
\end{tabular}
\caption{SAC-BC-N task specific BC hyperparameters.  Note the BC parameters are fixed within each task, i.e. do not vary based on dataset}
\label{SAC-BC-N-Spec}
\end{center}
\end{table}

\subsubsection*{Hardware}
The large scale experiment featured in Section \ref{implementation} was conducted on a machine with Intel Xeon E5-2698 v4 CPU, 512GB RAM and 8x Tesla V100-SXM2 32GB GPUs

\noindent
Experiments featured in Sections \ref{performance} and \ref{Expts_OnlineFT} were conducted on a machine with Intel Core i9 9900K CPU, 64GB RAM and 2x NVIDIA GeForce RTX 2080Ti 11GB TURBO GPUs.

\subsection*{Additional experimental results}
Following on from Section \ref{Expts_D4RL}, in Table \ref{tbl_adroit} we provide results for the full set of tasks from the Adroit domain using TD3-BC-N ($N=10, \beta=10$).  As with other approaches, we are only able to attain notable performance on the ``pen'' task.

\begin{table}[]
\begin{adjustbox}{width=\textwidth}
\begin{tabular}{|l||cccc|c|}
\hline
\textbf{Task / data set} & \textbf{CQL} & \textbf{IQL} & \textbf{EDAC} & \textbf{SAC-N} & \textbf{TD3-BC-N} \\ \hline
pen-cloned               & 39.2         & 37.3         & 68.2          & 64.1           & 67.2              \\ \
hammer-cloned            & 2.1          & 2.1          & 0.3           & 0.2            & 1.5               \\ \
door-cloned              & 0.4          & 1.6          & 9.6           & -0.3           & 0.0               \\ \
relocate-cloned          & -0.1         & -0.2         & 0             & 0              & 0.0               \\ \
pen-human                & 37.5         & 71.5         & 52.1          & 9.5            & 72.8              \\ 
hammer-human             & 4.4          & 1.4          & 0.8           & 0.3            & 0.8               \\ 
door-human               & 9.9          & 4.3          & 10.7          & -0.3           & 0                 \\ 
relocate-human           & 0.2          & 0.1          & 0.1           & -0.1           & -0.1              \\ \hline
average                  & 11.7         & 14.8         & 17.7          & 9.2            & 17.8              \\ \hline
\end{tabular}
\end{adjustbox}
    \caption{Performance comparison across Adroit benchmark.  Figures are normalised scores, with 0 and 100 representing random and expert policies, respectively.  As with other methods, our approach only achieves notable performance in the ``pen'' task}\label{tbl_adroit}
\end{table}

\noindent
Followong on from Section \ref{performance}, in Table \ref{tbl_app_mujoco} we provide results for MuJoCo tasks allowing the value of $\beta$ to vary within each task, observing a slight increase in performance.

\begin{table}[]
\begin{adjustbox}{width=\textwidth}
\begin{tabular}{|l||cc|ccc|ccc|}
\hline
Task / data set & EDAC  & \begin{tabular}[c]{@{}l@{}}SAC\\ -N\end{tabular} & \begin{tabular}[c]{@{}l@{}}TD3\\ -BC-N\\ (fixed)\end{tabular} & \begin{tabular}[c]{@{}l@{}}TD3\\ -BC-N\\ (variable)\end{tabular} & \begin{tabular}[c]{@{}l@{}}TD3\\ -BC-N\\ $\beta$\end{tabular} & \begin{tabular}[c]{@{}l@{}}SAC\\ -BC-N\\ (fixed)\end{tabular} & \begin{tabular}[c]{@{}l@{}}SAC\\ -BC-N\\ (variable)\end{tabular} & \begin{tabular}[c]{@{}l@{}}SAC\\ -BC-N\\ $\beta$\end{tabular} \\ \hline
halfcheetah     &       &                                                  &                                                               &                                                                  &                                                               &                                                               &                                                                  &                                                               \\
-medium         & 65.9  & 67.5                                             & 63.3 $\pm{0.1}$                                                         & 66.9     $\pm{0.2}$                                                         & 0                                                             & 65.6       $\pm{0.2}$                                                    & 65.6      $\pm{0.2}$                                                        & 0                                                             \\
-medium-replay  & 61.3  & 63.9                                             & 55.3 $\pm{0.1}$                                                          & 62.0  $\pm{0.2}$                                                            & 0                                                             & 61.5   $\pm{0.1}$                                                        & 61.5    $\pm{0.1}$                                                          & 0                                                             \\
-medium-expert  & 106.3 & 107.1                                            & 101.7  $\pm{0.3}$                                                        & 101.7  $\pm{0.3}$                                                           & 0.04                                                          & 102.6   $\pm{0.5}$                                                       & 102.6  $\pm{0.5}$                                                           & 0                                                             \\
-expert         & 106.8 & 105.2                                            & 103.8   $\pm{0.5}$                                                       & 103.8   $\pm{0.5}$                                                          & 0.04                                                          & 105.3   $\pm{0.1}$                                                       & 105.3    $\pm{0.1}$                                                         & 0                                                             \\ \hline
hopper          &       &                                                  &                                                               &                                                                  &                                                               &                                                               &                                                                  &                                                               \\
-medium         & 101.6 & 100.3                                            & 101.5  $\pm{0.3}$                                                        & 103.2  $\pm{0.0}$                                                           & 0.01                                                          & 101.2     $\pm{0.1}$                                                     & 101.2    $\pm{0.1}$                                                         & 0.0025                                                        \\
-medium-replay  & 101.0 & 101.8                                            & 99.1   $\pm{0.1}$                                                        & 100.0  $\pm{0.1}$                                                           & 0.01                                                          & 100.8  $\pm{0.2}$                                                        & 103.5   $\pm{0.4}$                                                          & 0                                                             \\
-medium-expert  & 110.7 & 110.1                                            & 112.3    $\pm{0.0}$                                                      & 112.3   $\pm{0.0}$                                                          & 0.03                                                          & 111.3     $\pm{0.1}$                                                     & 111.3  $\pm{0.1}$                                                           & 0.0025                                                        \\
-expert         & 110.1 & 110.3                                            & 112.7    $\pm{0.1}$                                                      & 112.7 $\pm{0.1}$                                                            & 0.03                                                          & 111.5     $\pm{0.1}$                                                     & 111.5 $\pm{0.1}$                                                            & 0.0025                                                        \\ \hline
walker2d        &       &                                                  &                                                               &                                                                  &                                                               &                                                               &                                                                  &                                                               \\
-medium         & 92.5  & 87.9                                             & 90.9     $\pm{0.2}$                                                      & 96.6  $\pm{0.3}$                                                            & 0.01                                                          & 85.3     $\pm{0.1}$                                                      & 92.1   $\pm{1.9}$                                                           & 0.001                                                         \\
-medium-replay  & 87.1  & 78.7                                             & 91.4    $\pm{0.4}$                                                       & 91.4   $\pm{0.4}$                                                           & 0.03                                                          & 90.8       $\pm{0.2}$                                                    & 96.6  $\pm{0.3}$                                                            & 0                                                             \\
-medium-expert  & 114.7 & 116.7                                            & 113.5   $\pm{0.1}$                                                       & 115.7   $\pm{0.3}$                                                          & 0                                                             & 110.9     $\pm{0.0}$                                                     & 117.5   $\pm{0.4}$                                                          & 0.001                                                         \\
-expert         & 115.1 & 107.4                                            & 113.2    $\pm{0.0}$                                                      & 113.2    $\pm{0.0}$                                                         & 0.03                                                          & 110.4    $\pm{0.0}$                                                      & 110.4 $\pm{0.0}$                                                            & 0.0025                                                        \\ \hline
mujoco average  & 97.8  & 96.4                                             & 96.6                                                          & 98.3                                                             &                                                               & 96.4                                                          & 98.3                                                             &                                                               \\ \hline
\end{tabular}
\end{adjustbox}
    \caption{Performance comparison across MuJoCo benchmark, allowing $\beta$ to vary within each task.  Figures are normalised scores, with 0 and 100 representing random and expert policies, respectively.  Allowing $\beta$ to vary marginally enhances performance}\label{tbl_app_mujoco}
\end{table}

\subsection*{Further details regarding computational efficiency experiments}
To ensure a fair comparison of computational efficiency, we implement our own versions of baselines (available in our code repository) based on author published source code and the CORL repository \citep{tarasov2022corl}, and run them on the same hardware/software configuration.  In terms of hardware we use a machine with a Intel Core i9 9900K CPU, 64GB RAM and 2x NVIDIA GeForce RTX 2080Ti 11GB TURBO GPUs.  In terms of software we use PyTorch (version 1.9.1+cu102).

The ensemble architecture for TD3-BC-N, SAC-BC-N, SAC-N, EDAC and MSG is exactly the same.  Each Q-network comprises a 3-layer MLP with ReLU activation functions and 256 nodes, taking as input a state-action pair and outputting a Q-value.  For TD3-BC-N the policy network comprises a 3-layer MLP with ReLU activation functions and 256 nodes, taking as input a state and outputting an action bound to [-1, 1] via tanh transformation.  For SAC-BC-N, SAC-N, EDAC and MSG the policy network comprises the same architecture but instead outputs the mean and standard deviation of a Gaussian distribution which is also bound to [-1, 1] via tanh transformation.  

For CQL, we use a dual critic, with each Q-network comprising a 3-layer MLP with ReLU activation functions and 256 nodes, taking as input a state-action pair and outputting a Q-value.  The policy network comprises a 3-layer MLP with ReLU activation functions and 256 nodes outputting the mean and standard deviation of a Gaussian distribution which is bound to [-1, 1] via tanh transformation.

For IQL, we use a dual critic, with each Q-network comprising a 2-layer MLP with ReLU activation functions and 256 nodes, taking as input a state-action pair and outputting a Q-value.  We use a single state-value network comprising a 2-layer MLP with ReLU activation functions and 256 nodes, taking as input a state and outputting a state-value.  The policy network comprises a 2-layer MLP with ReLU activation functions and 256 nodes outputting a tanh transformed mean and standard deviation of a Gaussian distribution.

For TD3-BC, we use a dual critic, with each Q-network comprising a 2-layer MLP with ReLU activation functions and 256 nodes, taking as input a state-action pair and outputting a Q-value.  The policy network comprises a 2-layer MLP with ReLU activation functions and 256 nodes, taking as input a state and outputting an action bound to [-1, 1] via tanh transformation

For all algorithms we use the Adam optimiser \citep{kingma2014adam} and a batch size of 256.

For each algorithm, we record the training time for 10,000 gradient steps and scale by the total number of gradient steps to arrive at the total computation time.  We detail these calculations in Table in \ref{tbl_comp_calc}.

\begin{table}[]
\begin{adjustbox}{width=\textwidth}
\begin{tabular}{|lcccc|}
\hline
\multicolumn{1}{|l}{\textbf{Algorithm}} & \textbf{\begin{tabular}[c]{@{}c@{}}Runtime \\ (sec/epoch*)\end{tabular}} & \textbf{\begin{tabular}[c]{@{}c@{}}Total gradient\\ steps\end{tabular}} & \textbf{\begin{tabular}[c]{@{}c@{}}Total runtime\\ (hours)\end{tabular}} & \multicolumn{1}{c|}{\textbf{\begin{tabular}[c]{@{}c@{}}GPU memory \\ (GB)\end{tabular}}} \\ \hline
SAC-10                                  & 60                                                                       & 3M                                                                      & 5.0                                                                      & 1.2                                                                                      \\
SAC-20                                  & 61                                                                       & 3M                                                                      & 5.1                                                                      & 1.3                                                                                      \\
SAC-100                                 & 142                                                                      & 3M                                                                      & 11.8                                                                     & 1.6                                                                                      \\
SAC-200                                 & 251                                                                      & 3M                                                                      & 20.9                                                                     & 2.0                                                                                      \\
SAC-500                                 & 588                                                                      & 3M                                                                      & 49.0                                                                     & 3.5                                                                                      \\
EDAC-10                                 & 76                                                                       & 3M                                                                      & 6.3                                                                      & 1.2                                                                                      \\
EDAC-20                                 & 86                                                                       & 3M                                                                      & 7.2                                                                      & 1.3                                                                                      \\
EDAC-50                                 & 136                                                                      & 3M                                                                      & 11.3                                                                     & 1.5                                                                                      \\
MSG-4                                   & 65                                                                       & 2M                                                                      & 3.6                                                                      & 1.2                                                                                      \\
MSG-64                                  & 131                                                                      & 2M                                                                      & 7.3                                                                      & 1.5                                                                                      \\
CQL                                     & 123                                                                      & 1M                                                                      & 3.4                                                                      & 1.3                                                                                      \\
IQL                                     & 54                                                                       & 1M                                                                      & 1.5                                                                      & 1.2                                                                                      \\
TD3-BC                                  & 26                                                                       & 1M                                                                      & 0.7                                                                      & 1.2                                                                                      \\
SAC-BC-10                               & 62                                                                       & 1M                                                                      & 1.7                                                                      & 1.2                                                                                      \\
TD3-BC-10                               & 31                                                                       & 1M                                                                      & 0.9                                                                      & 1.2                                                                                      \\ \hline
\end{tabular}
\end{adjustbox}
    \caption{Computation time calculation details.  *1 epoch=10,000 gradient steps}\label{tbl_comp_calc}
\end{table}

\subsection*{Additional plots}\label{AppPlots}
\begin{figure}[] % h=here, t=top, b=bottom, p=new page, !=other preferences
            \begin{center}
                \includegraphics[width=1\textwidth]{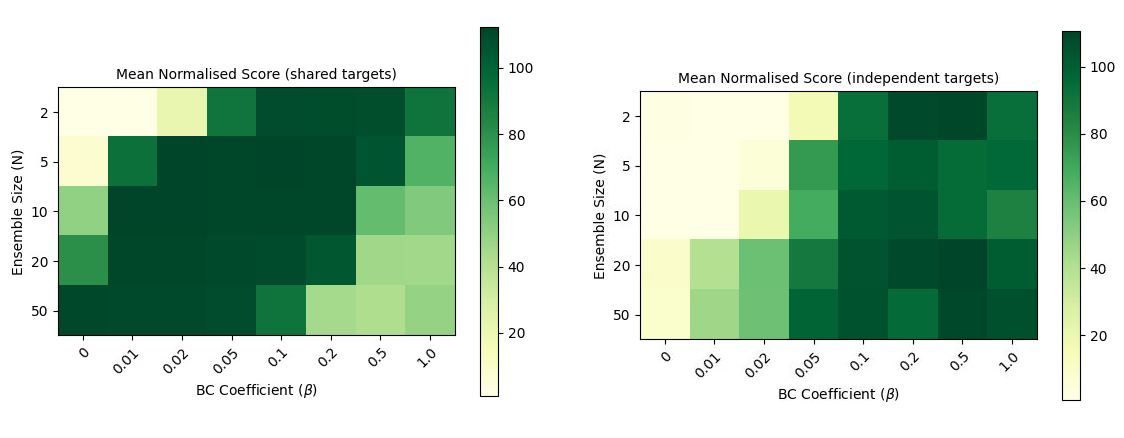}
            \end{center}
            \caption{Performance as a function of $N$ and $\beta$.  Lower values of $\beta$ require larger values of $N$ and smaller values of $N$ require higher values of $\beta$.  Shared targets (left) - If both $N$ and $\beta$ are large, the uncertainty in Q-value estimates for OOD actions is too high, and thus the penalty applied too severe, leading the agent to prefer actions similar to those of the data.  Independent targets (right) - the decline in performance for large $N$ and $\beta$ is not observed but this may be a result of both values needing to be higher in general, and hence for even larger $N$ and $\beta$ this outcomes may also be observed }
            \label{fig:AppPer}
        \end{figure}

\begin{figure}[] % h=here, t=top, b=bottom, p=new page, !=other preferences
            \begin{center}
                \includegraphics[width=1\textwidth]{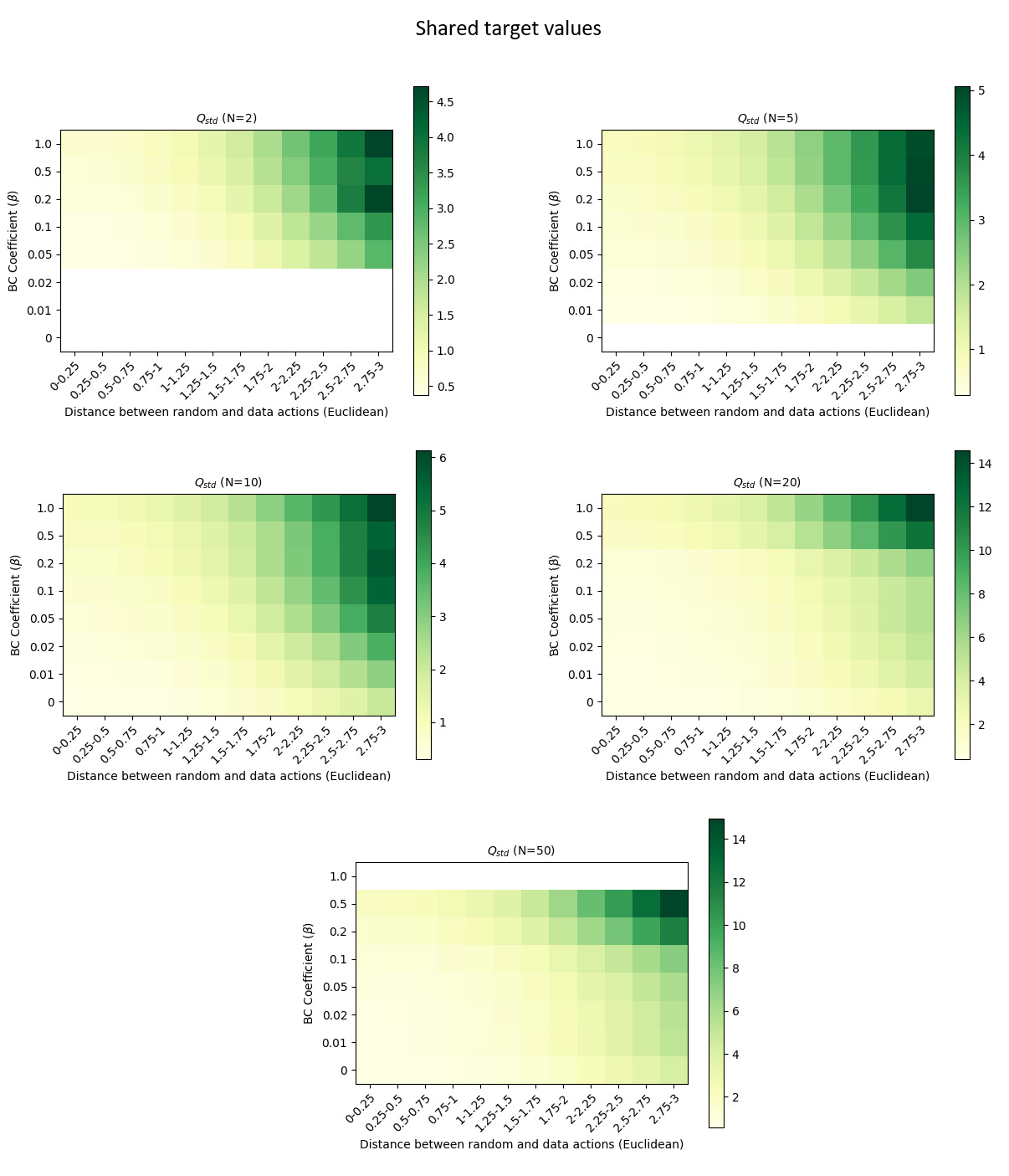}
            \end{center}
            \caption{Standard deviation as a function of distance, $N$ and $\beta$ (shared target values).  As the distance between random and data actions increases so too does the level of uncertainty, becoming more pronounced as $\beta$ and $N$ get larger.  White space is used to represent erroneous values due to unreliable Q-values estimates resulting from divergent critic loss during training}
            \label{fig:AppStd}
        \end{figure}

\begin{figure}[] % h=here, t=top, b=bottom, p=new page, !=other preferences
            \begin{center}
                \includegraphics[width=1\textwidth]{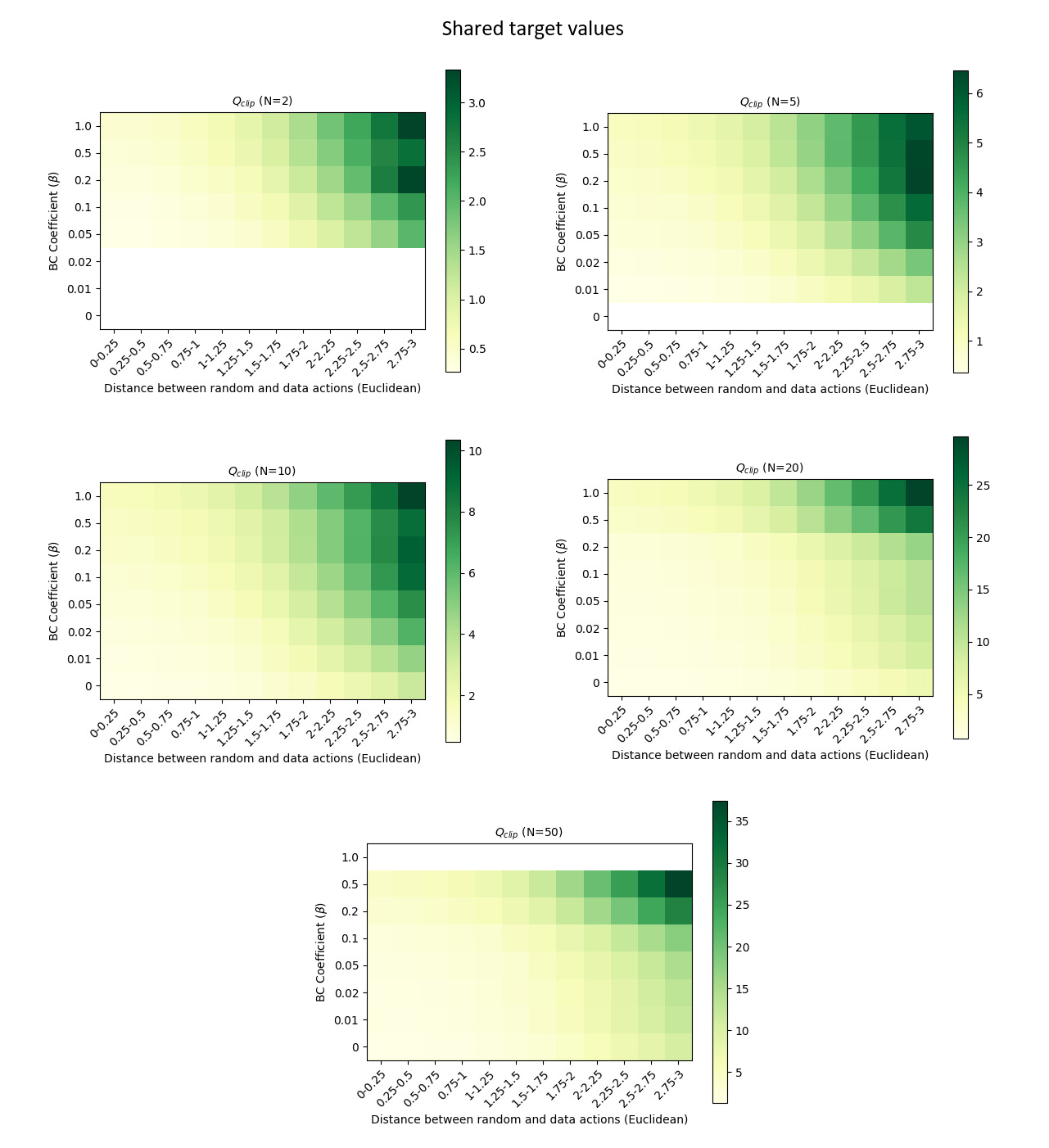}
            \end{center}
            \caption{Clip penalty as a function of distance, $N$ and $\beta$ (shared target values).  As the distance between random and data actions increases so too does the level of uncertainty, becoming more pronounced as $\beta$ and $N$ get larger.  White space is used to represent erroneous values due to unreliable Q-values estimates resulting from divergent critic loss during training}
            \label{fig:AppClipPen}
        \end{figure}

\begin{figure}[] % h=here, t=top, b=bottom, p=new page, !=other preferences
            \begin{center}
                \includegraphics[width=1\textwidth]{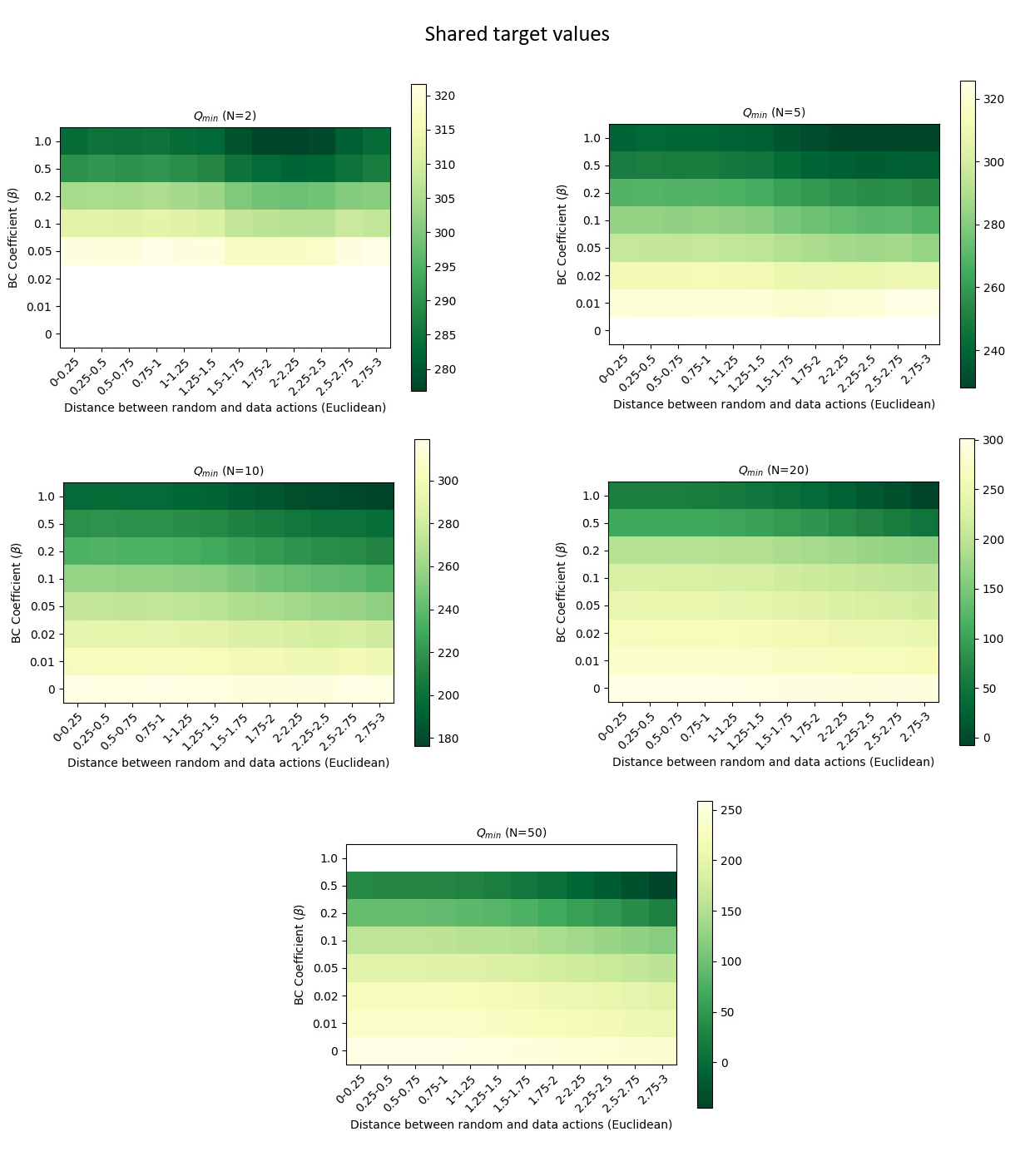}
            \end{center}
            \caption{$Q_{min}$ as a function of distance, N and $\beta$ (shared target values).  As the distance between random and data actions increases, $Q_{min}$ decreases, with this decrease more pronounced as $\beta$ and $N$ get larger.  White space is used to represent erroneous values due to unreliable Q-values estimates resulting from divergent critic loss during training}
            \label{fig:AppMin}
        \end{figure}

\begin{figure}[] % h=here, t=top, b=bottom, p=new page, !=other preferences
            \begin{center}
                \includegraphics[width=1\textwidth]{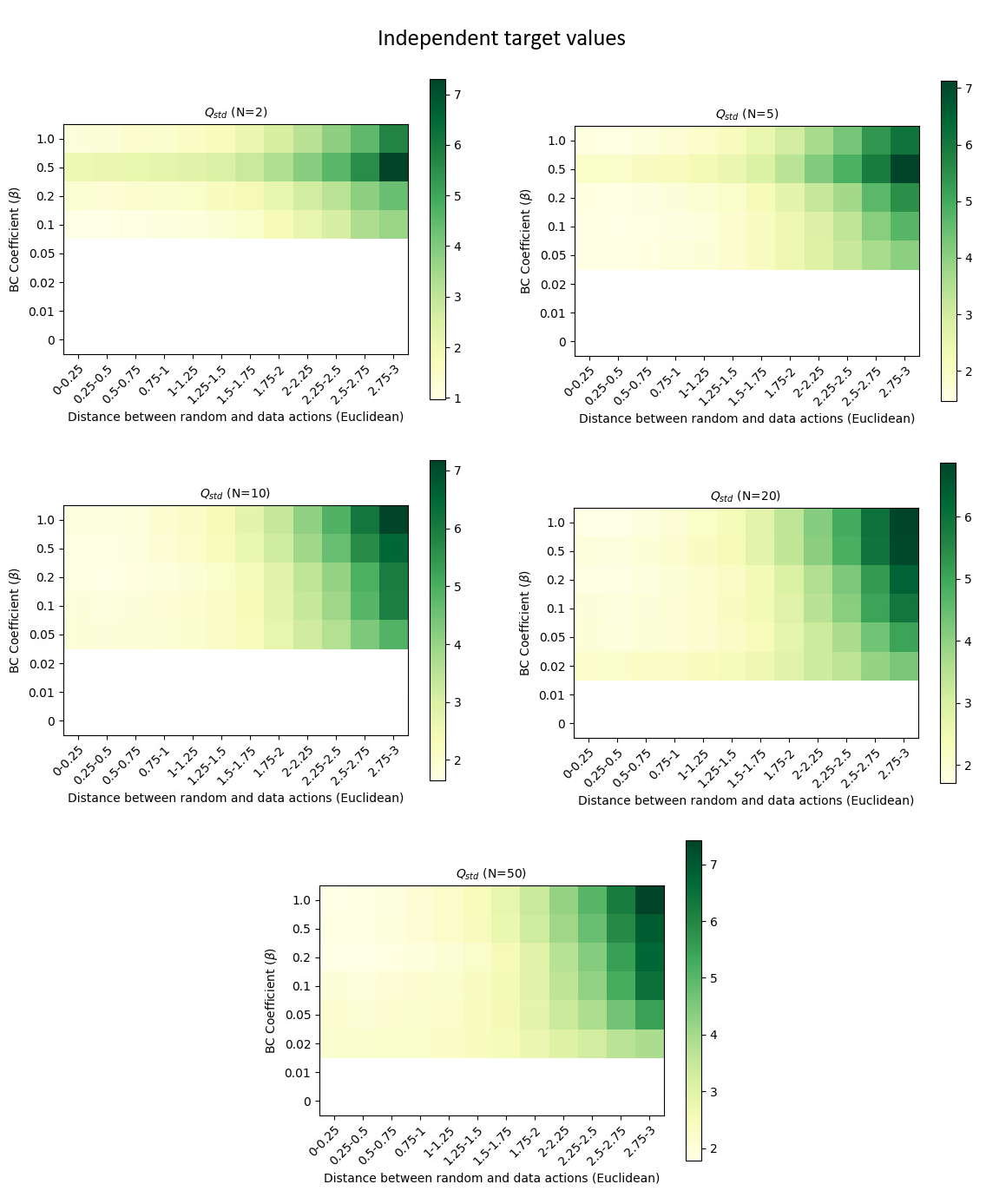}
            \end{center}
            \caption{Standard deviation as a function of distance, $N$ and $\beta$ (independent target values).  As the distance between random and data actions increases so too does the level of uncertainty, becoming more pronounced as $\beta$ and $N$ get larger.  White space is used to represent erroneous values due to unreliable Q-values estimates resulting from divergent critic loss during training}
            \label{fig:IndepAppStd}
        \end{figure}

\begin{figure}[] % h=here, t=top, b=bottom, p=new page, !=other preferences
            \begin{center}
                \includegraphics[width=1\textwidth]{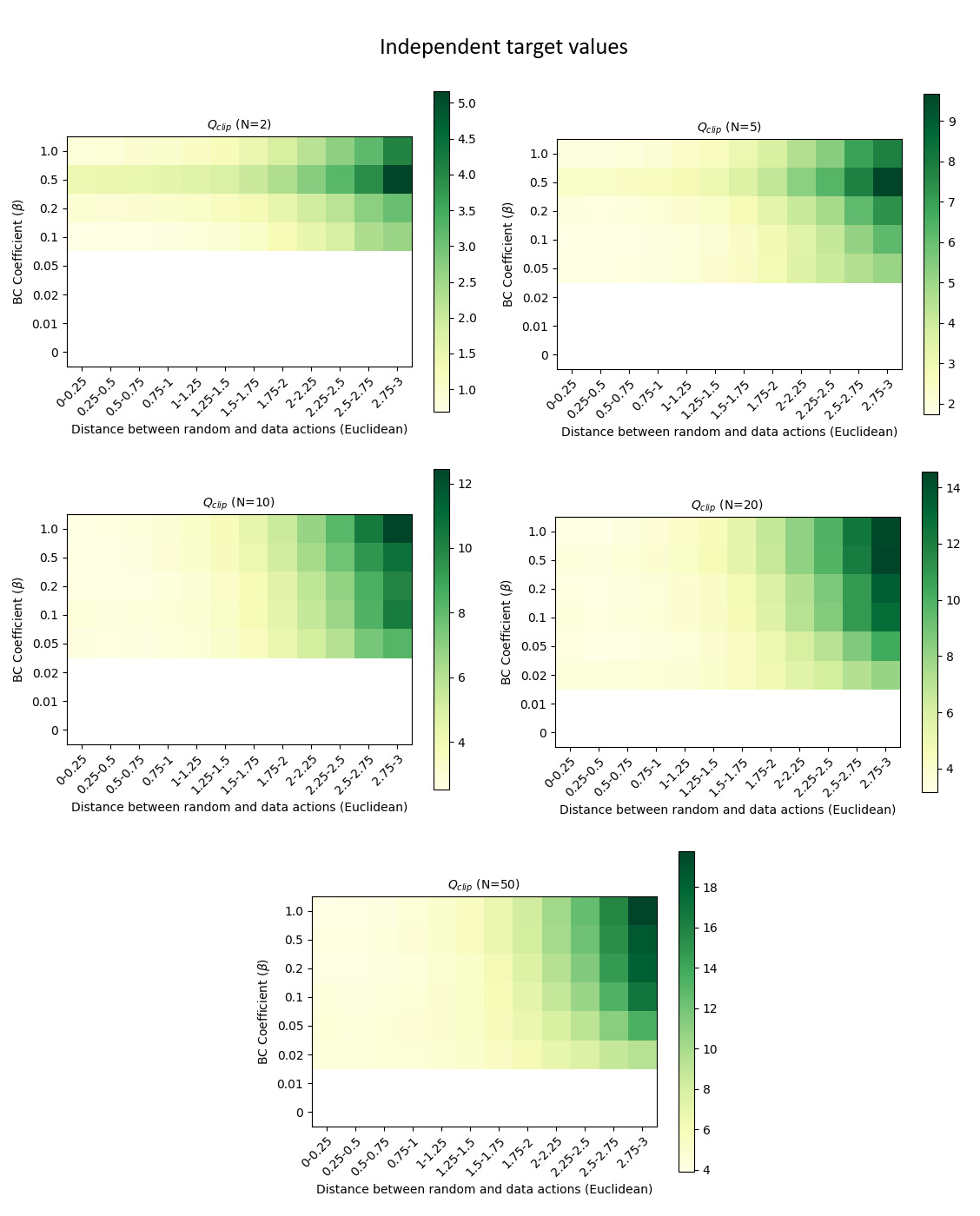}
            \end{center}
            \caption{Clip penalty as a function of distance, $N$ and $\beta$ (shared independent values).  As the distance between random and data actions increases so too does the level of uncertainty, becoming more pronounced as $\beta$ and $N$ get larger.  White space is used to represent erroneous values due to unreliable Q-values estimates resulting from divergent critic loss during training}
            \label{fig:IndepAppClipPen}
        \end{figure}

\begin{figure}[] % h=here, t=top, b=bottom, p=new page, !=other preferences
            \begin{center}
                \includegraphics[width=1\textwidth]{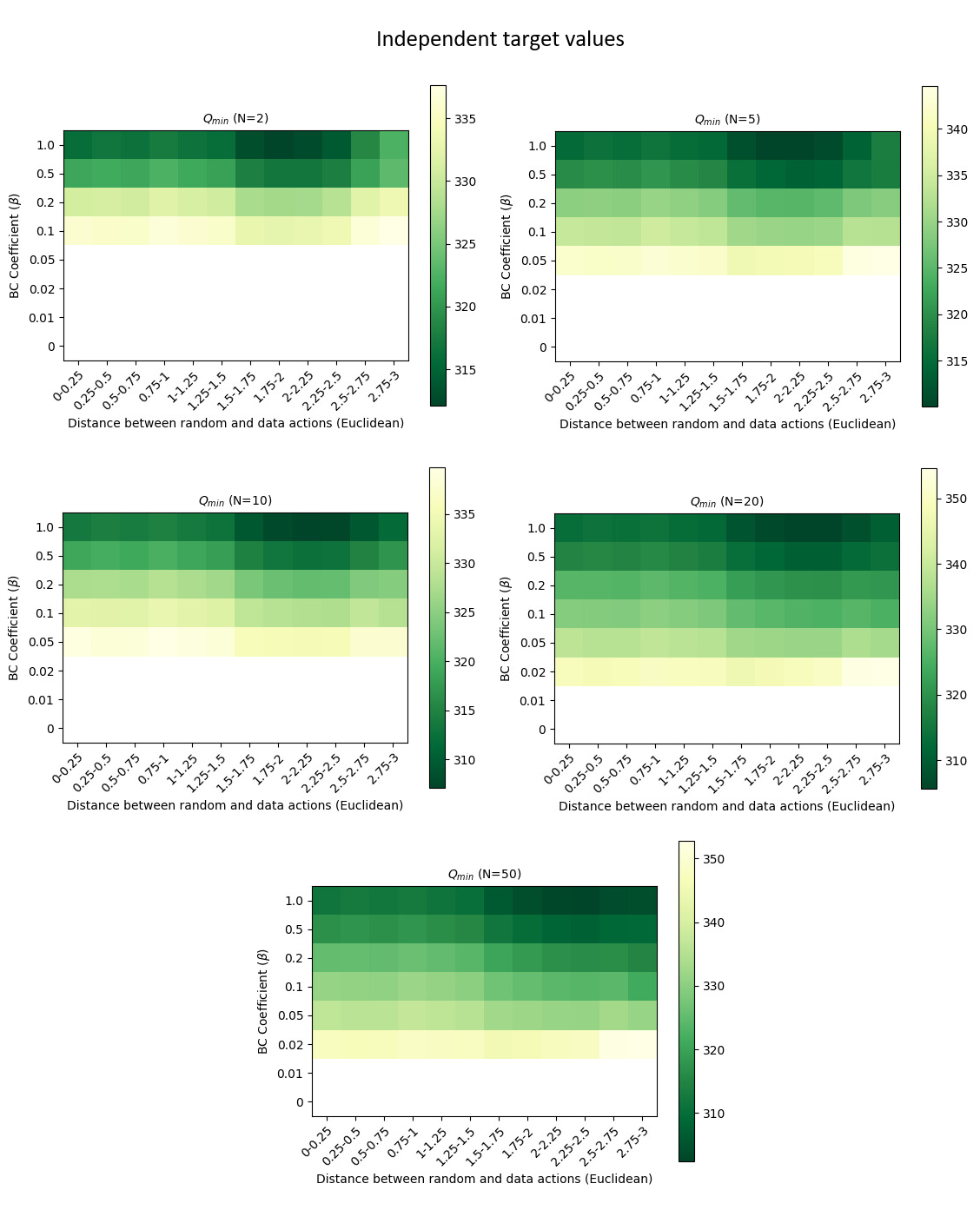}
            \end{center}
            \caption{$Q_{min}$ as a function of distance, N and $\beta$ (independent target values).  As the distance between random and data actions increases, $Q_{min}$ decreases, with this decrease more pronounced as $\beta$ and $N$ get larger.  White space is used to represent erroneous values due to unreliable Q-values estimates resulting from divergent critic loss during training}
            \label{fig:IndepAppMin}
        \end{figure}

Following on from Section \ref{interpretation}, we provide the complete set of plots from our case study using the ``hopper-medium-expert'' dataset.  Figure \ref{fig:AppPer} summarises performance for shared and independent target values, with Figures \ref{fig:AppStd}-\ref{fig:AppMin} and \ref{fig:IndepAppStd}-\ref{fig:IndepAppMin} showing $Q_{std}$, $Q_{clip}$ and $Q_{min}$ for shared and independent target values, respectively.

\begin{figure}[] % h=here, t=top, b=bottom, p=new page, !=other preferences
            \begin{center}
                \includegraphics[width=1\textwidth]{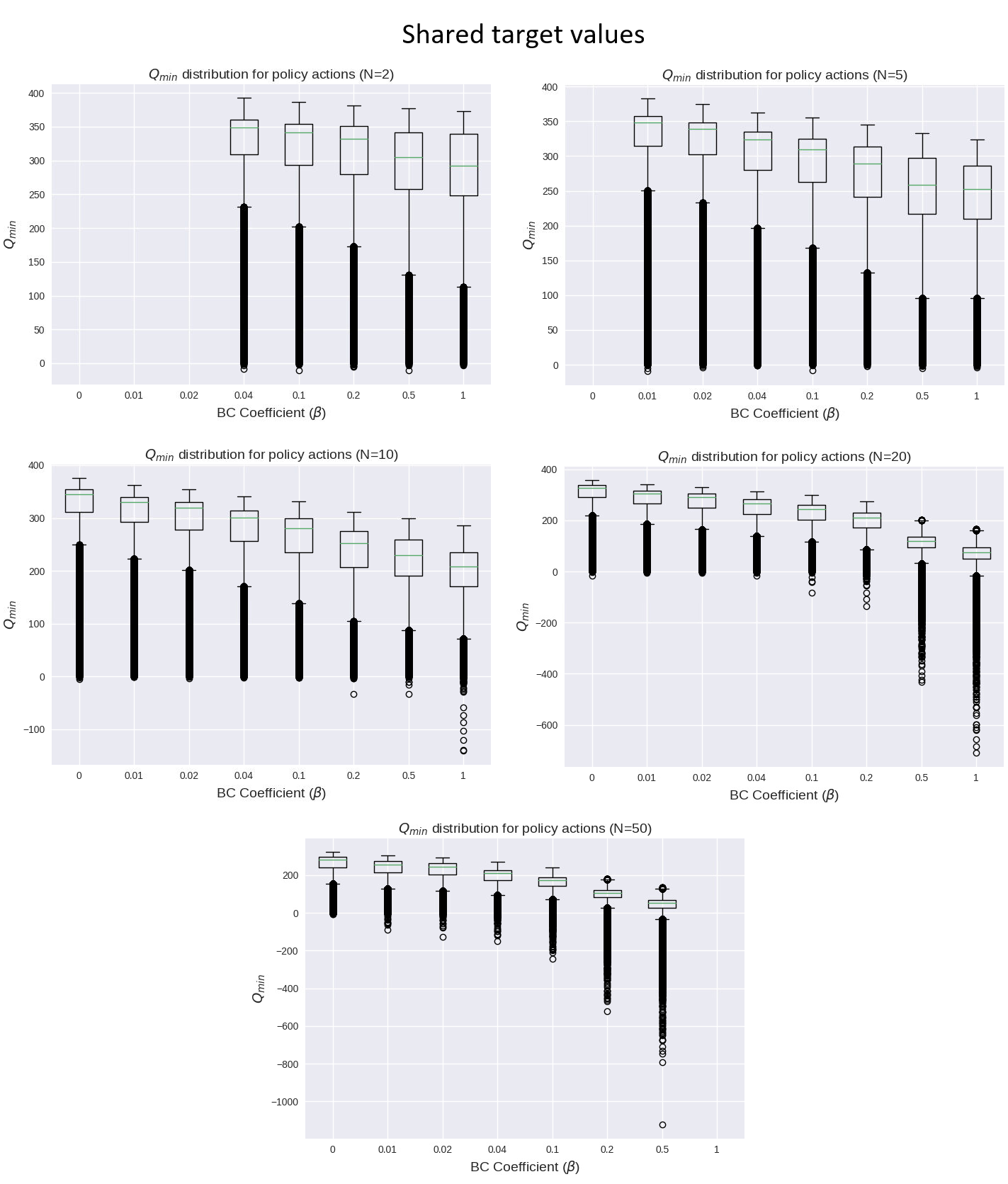}
            \end{center}
            \caption{Distribution of $Q_{min}$ for policy actions (shared target values).  In general, the higher the value of $\beta$ the lower the values of $Q_{min}$, as Q-value estimates are penalised more heavily.  This is particularly noticeable when $N$ and $\beta$ are large, contributing to declining performance as observed in Figure 10}
            \label{fig:BoxMin}
        \end{figure}

\begin{figure}[] % h=here, t=top, b=bottom, p=new page, !=other preferences
            \begin{center}
                \includegraphics[width=1\textwidth]{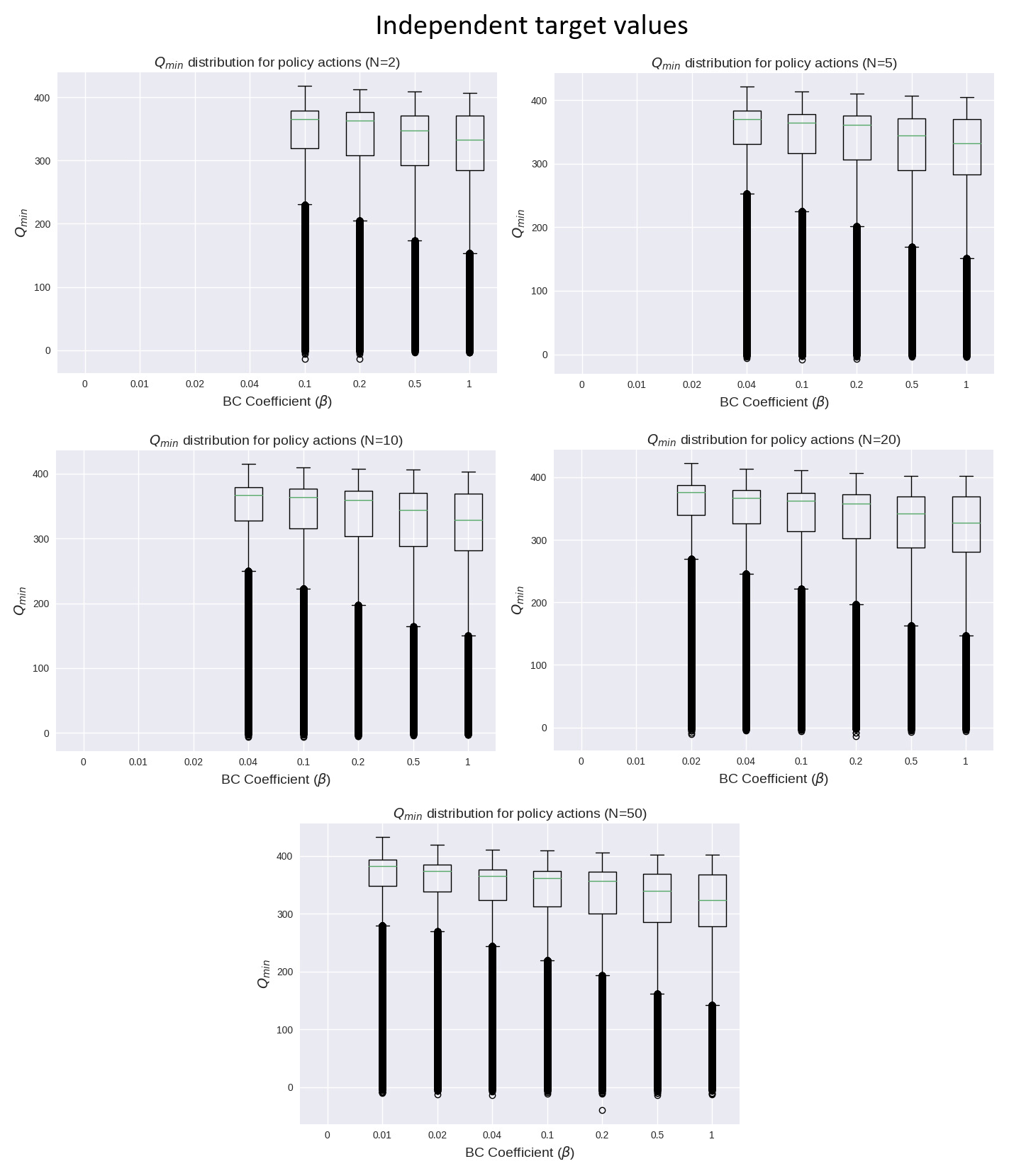}
            \end{center}
            \caption{Distribution of $Q_{min}$ for policy actions (independent target values).  In general, the higher the value of $\beta$ the lower the values of $Q_{min}$, as Q-value estimates are penalised more heavily.  For this range of $N$ and $\beta$ the distributions do not exhibit extreme estimates as in Figure 17, consistent with performance as observed in Figure 10.  However, this may be the case for higher $N$ and $\beta$.}
            \label{fig:IndepBoxMin}
        \end{figure}

\begin{figure}[] % h=here, t=top, b=bottom, p=new page, !=other preferences
            \begin{center}
                \includegraphics[width=1\textwidth]{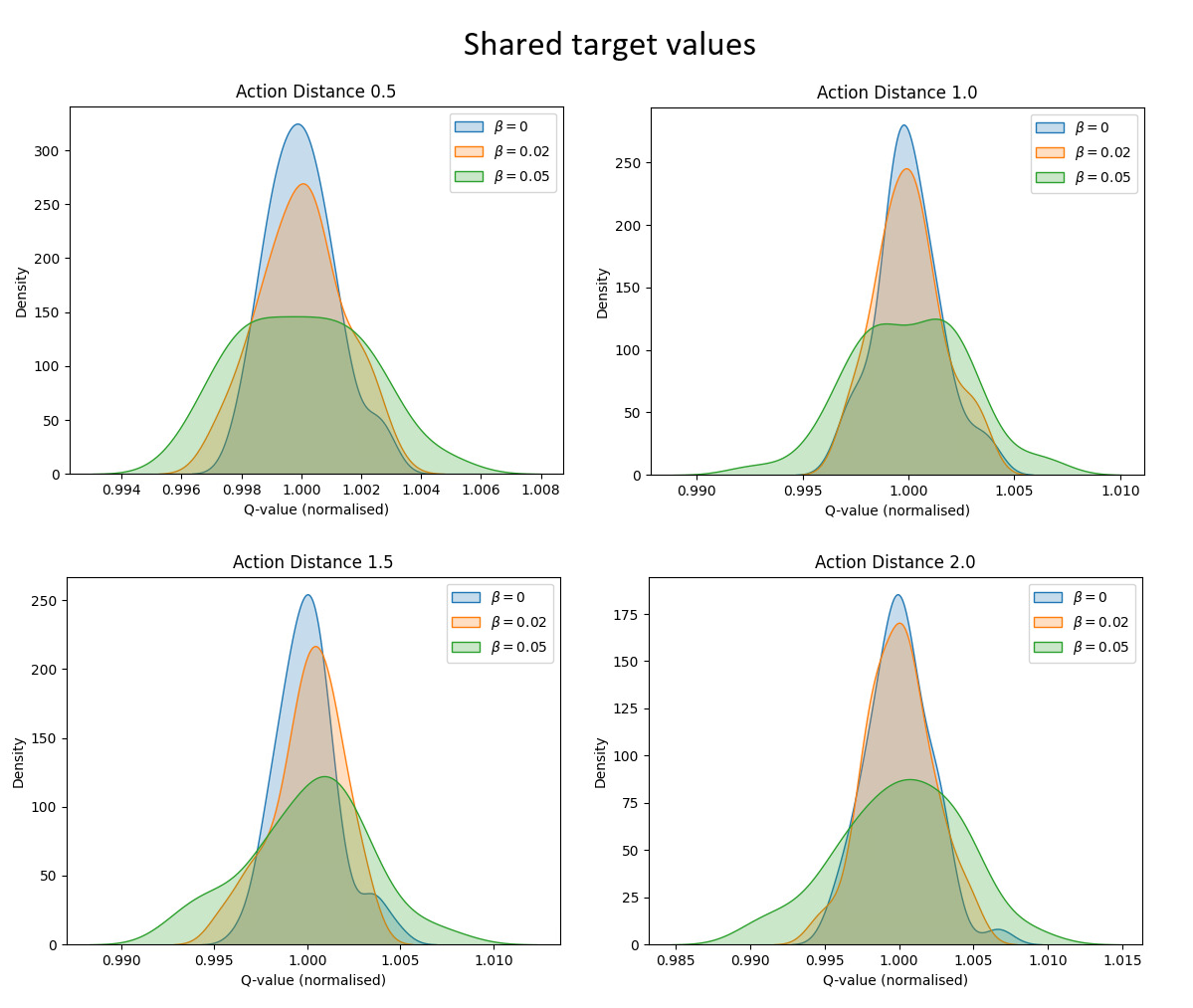}
            \end{center}
            \caption{Examples density estimates of Q-functions (shared target values, $N=50$).  Q-values are normalised to allow easier comparison of uncertainty.  As $\beta$ increases so too does the variance in Q-value estimates}
            \label{fig:QKDE}
        \end{figure}

\begin{figure}[] % h=here, t=top, b=bottom, p=new page, !=other preferences
            \begin{center}
                \includegraphics[width=1\textwidth]{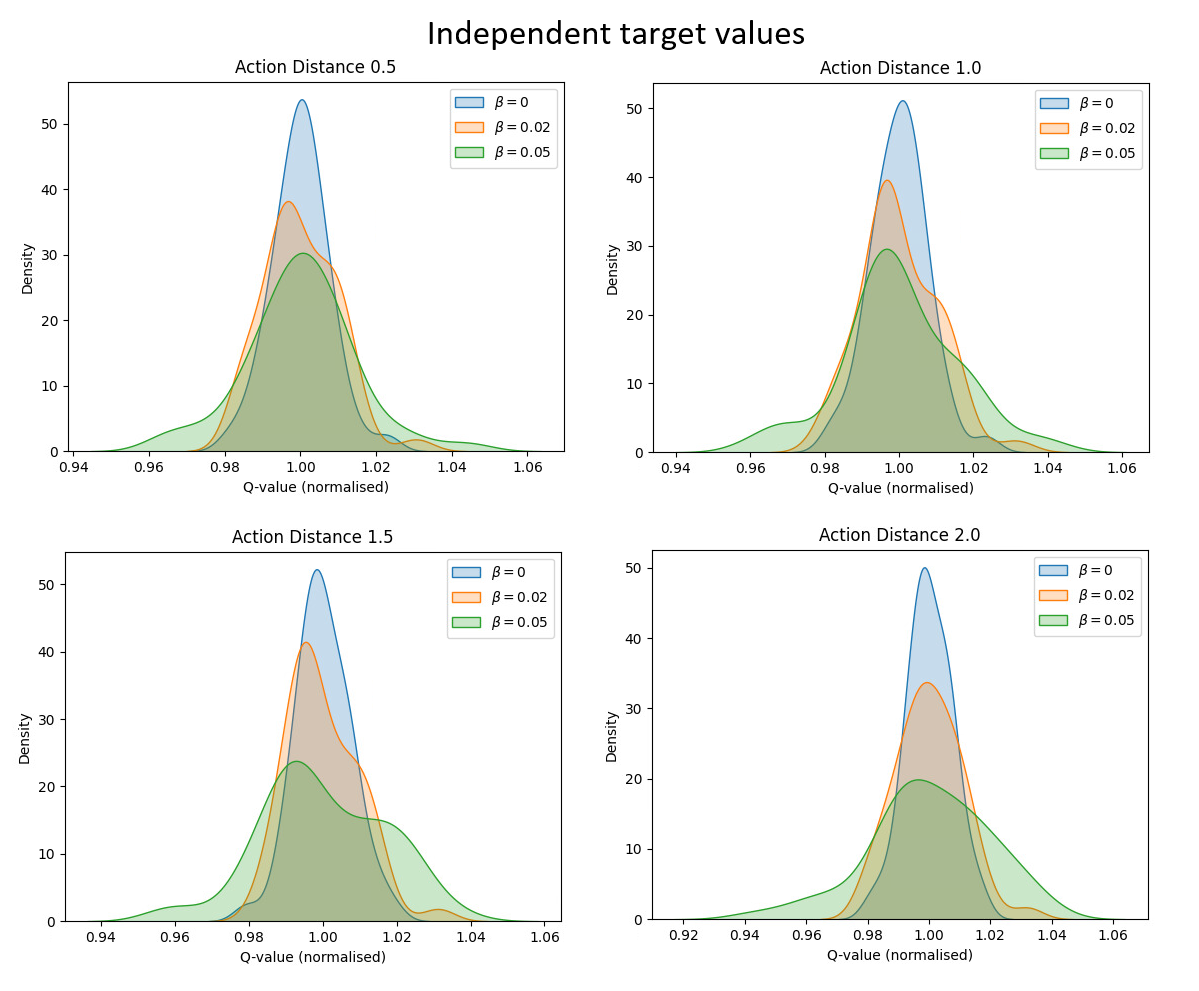}
            \end{center}
            \caption{Examples density estimates of Q-functions (independent target values, $N=50$).  Q-values are normalised to allow easier comparison of uncertainty.  As $\beta$ increases so too does the variance in Q-value estimates}
            \label{fig:IndepQKDE}
        \end{figure}

We also provide plots examining the distribution of $Q_{min}$ for policy actions in Figures \ref{fig:BoxMin} and \ref{fig:IndepBoxMin}, and examples density estimates of Q-value distributions for individual state-action pairs in Figures \ref{fig:QKDE} and \ref{fig:IndepQKDE}.  To allow for better estimates of density, we use ensembles of size $N=50$, and to allow easier comparisons of uncertainty we normalise Q-values by dividing by the mean of the absolute value across the ensemble (similar to Section \ref{PCUE}).  Note this normalisation only changes the location of the distribution, not the variance.

\end{document}